\documentclass{article} 
\usepackage{iclr2027_conference,times}

\usepackage{amsmath,amsfonts,bm}

\def\eqref#1{equation~\ref{#1}}

\def\1{\bm{1}}

\DeclareMathAlphabet{\mathsfit}{\encodingdefault}{\sfdefault}{m}{sl}
\SetMathAlphabet{\mathsfit}{bold}{\encodingdefault}{\sfdefault}{bx}{n}

\usepackage{graphicx}
\usepackage{hyperref}
\usepackage{url}
\usepackage{booktabs}
\usepackage{multirow}
\usepackage{graphicx}
\usepackage{amssymb}
\usepackage[table]{xcolor}
\definecolor{vdTableHead}{RGB}{231,237,241}
\definecolor{vdTableGroup}{RGB}{242,245,247}
\definecolor{vdTableOursBand}{RGB}{222,235,242}
\definecolor{vdTableOursRow}{RGB}{239,246,249}

\title{VD-DeepStack: Bridging Visual Comparison and Language Reasoning for Few-Shot Anomaly Detection}

\author{%
  \normalfont
  \begin{tabular}[t]{@{}c@{}}
    Mengyang Zhao\textsuperscript{1} \quad Zhuolin He\textsuperscript{1} \quad Haiyang Yu\textsuperscript{1} \quad Yuxuan Liang\textsuperscript{1} \\[2pt]
    Yifang Xu\textsuperscript{1} \quad Yuchuan Wu\textsuperscript{1} \quad Xiaolei Chen\textsuperscript{1} \quad Zhengtao Yao\textsuperscript{3} \\[2pt]
    Fan Shi\textsuperscript{4} \quad Yang Liu\textsuperscript{2} \quad Bin Li\textsuperscript{1} \quad Xiangyang Xue\textsuperscript{1} \\[6pt]
    \small \textsuperscript{1}Fudan University \quad \textsuperscript{2}Tongji University \\[1pt]
    \small \textsuperscript{3}University of Southern California \quad \textsuperscript{4}Fuzhou University  
  \end{tabular}%
}

\iclrfinalcopy
\hypersetup{%
  pdftitle={VD-DeepStack: Bridging Visual Comparison and Language Reasoning for Few-Shot Anomaly Detection},
  pdfauthor={Mengyang Zhao, Zhuolin He, Haiyang Yu, Yuxuan Liang, Yifang Xu, Yuchuan Wu, Xiaolei Chen, Zhengtao Yao, Fan Shi, Yang Liu, Bin Li, Xiangyang Xue}
}
\begin{document}

\begingroup
\centering
\maketitle
\endgroup
\pagestyle{plain}
\thispagestyle{plain}

\begin{abstract}

Few-shot visual anomaly detection is fundamentally a visual comparison task, requiring fine-grained inspection of a query against normal references. Many recent methods based on large vision-language models (LVLMs) emphasize comparative reasoning through language chain-of-thought. Yet discrete, abstract descriptions may underrepresent dense, fine-grained visual differences, leaving a gap between visual comparison and its expression in language. To address this gap, we propose Visual Difference DeepStack (VD-DeepStack), which explicitly conditions language reasoning on query–reference visual differences. Specifically, we fuse DINO features with the LVLM visual hierarchy to strengthen fine-grained representations, then construct dense difference evidence from residuals between query features and softly matched reference features. 
The difference-evidence path injects spatially weighted difference vectors into query-image states at multiple decoder depths, while an auxiliary visual-context path provides fine-grained appearance information to support their interpretation. Experiments on 4 industrial and 2 medical anomaly benchmarks demonstrate substantial improvements in few-shot anomaly detection over baselines relying on textual comparative reasoning. These results support mitigating the visual comparison–reasoning gap through the joint design of comparison representations and their integration into the decoder. Code will be released upon acceptance.

\end{abstract}

\section{Introduction}
\label{sec:introduction}

Reference-based few-shot visual anomaly detection is a \emph{comparison-first} task: a model must identify task-relevant deviations from normal appearance using only a few normal references~\citep{huang2022regad,zhu2024inctrl}. Whether a local mark is a defect depends on the expected texture and structure, not merely the object's semantic category. Large vision-language models (LVLMs) extend this task beyond anomaly scoring to natural-language reasoning and interpretation. This raises a central question: \emph{how should visual comparison enter the reasoning process?}

\begin{figure*}[tbhp]
\centering
\includegraphics[width=0.995\textwidth]{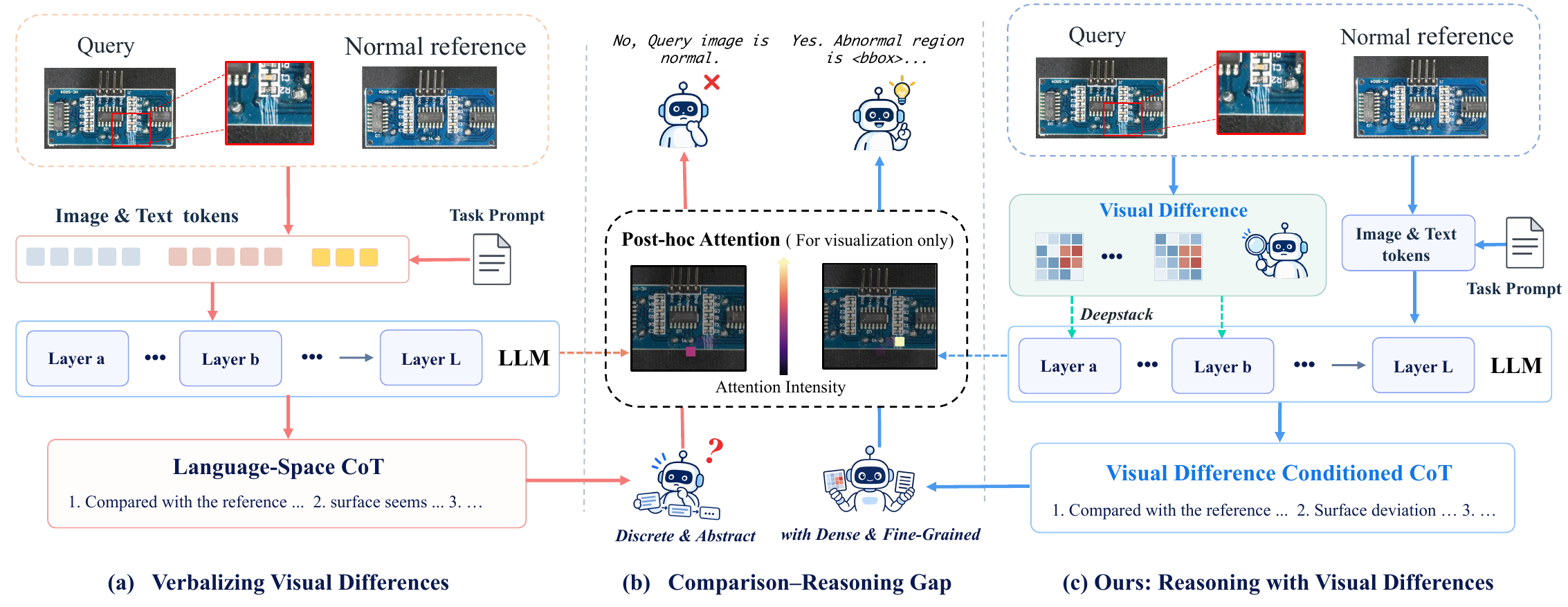}
\caption{\textbf{Verbal comparison versus visual-difference conditioning.}
(a,c) Two reasoning routes using the same query--reference pair.
(b) Example predictions and post-hoc attention maps, shown for interpretation only and not used as model inputs.}
\label{fig:intro}
\end{figure*}

Recent methods largely address this question through reasoning supervision. IAD-R1~\citep{li2026iadr1} strengthens anomaly-specific chain-of-thought (CoT) reasoning, while AD-FM~\citep{liao2026adfm} organizes inspection into multiple reasoning stages with localization-aware rewards. Comparison is also increasingly explicit: MMR-AD~\citep{yao2026mmrad} constructs comparative CoTs from query--reference pairs, and JUDO~\citep{kang2026judo} learns juxtaposed segmentation and domain-oriented reasoning. These advances demonstrate the value of comparative reasoning. However, richer descriptions of visual differences do not necessarily preserve the fine-grained information underlying those differences.

We characterize this mismatch as the \textbf{visual comparison--reasoning gap}. Language CoT expresses visual comparisons through discrete, semantically abstract descriptions, which may underrepresent the dense, fine-grained differences needed to distinguish anomalies from normal variation. Improving how a model describes these differences does not necessarily improve how precisely it compares local texture, structure, and appearance against a normal reference. This motivates a complementary direction: making fine-grained visual differences more directly available to language reasoning, alongside supervision on their verbal interpretation. As illustrated in Fig.~\ref{fig:intro}, our aim is to move beyond \emph{talking about differences} toward \emph{thinking with visual differences}, while retaining language for interpretation and explanation.

To address this gap, we propose \textbf{Visual Difference
DeepStack (VD-DeepStack)}. Its design is motivated by two
considerations. First, informative comparison requires
representations sensitive to subtle appearance differences
even when the query and reference share the same object
semantics. We therefore enhance the LVLM visual hierarchy
with DINO features and construct dense difference evidence
through soft query--reference matching. Second, we aim to
make these differences directly available to the language
decoder while retaining the appearance context that supports
their interpretation. Building on
DeepStack~\citep{meng2024deepstack}, depth-specific residual
writers inject spatially weighted difference vectors into
query-image states at multiple decoder depths. An auxiliary
visual-context path supplies fine-grained appearance information
to both images to support interpretation of the injected
evidence. These updates condition subsequent autoregressive
reasoning on fine-grained visual comparison without adding
evidence tokens.
Experiments on multiple visual anomaly benchmarks demonstrate substantial improvements in few-shot detection and localization over baselines relying on textual comparative reasoning. We further examine the contributions of the visual-context and difference-evidence paths through ablation studies. Comparisons with score-only and evidence-token variants evaluate alternative forms of visual conditioning.

Our contributions are threefold:
\begin{itemize}
\item We highlight a potential visual comparison--reasoning gap:
language-based descriptions may underrepresent fine-grained
visual differences. Motivated by this insight, we propose
\textbf{VD-DeepStack}, which constructs dense query--reference
difference representations to condition autoregressive anomaly
reasoning.
\item  We develop a DeepStack-based conditioning interface that
injects difference evidence into query-image states through
residual updates at multiple decoder depths. An auxiliary
visual-context path supplies fine-grained appearance features
to both reference and query states.
\item   We demonstrate substantial improvements over textual CoT
baselines on industrial anomaly benchmarks, with additional gains
on medical anomaly detection datasets supporting cross-domain
generalization. Ablation studies assess the contributions of
difference evidence and auxiliary visual context.
\end{itemize}

\section{Related Work}
\label{sec:related_work}

\paragraph{Zero- and few-shot visual anomaly detection.} Traditional few-shot approaches model normal appearance through feature registration~\citep{huang2022regad} or reconstruction of query features from normal supports~\citep{fang2023fastrecon}.
Recently, WinCLIP~\citep{jeong2023winclip} uses handcrafted prompt ensembles for zero-shot detection, while its few-shot extension, WinCLIP+, incorporates normal-reference matching.
PromptAD~\citep{li2024promptad} learns prompts using only normal samples for few-shot detection.
For zero-shot transfer, AnomalyCLIP and AdaCLIP learn object-agnostic and hybrid prompts, respectively, on auxiliary data~\citep{zhou2024anomalyclip,cao2024adaclip}.
FiLo~\citep{gu2024filo} incorporates fine-grained anomaly descriptions for zero-shot detection, while FiLo++~\citep{gu2026filopp} additionally supports few-shot detection through position-enhanced normal patch matching.
Among methods that explicitly model visual comparison, InCTRL~\citep{zhu2024inctrl} learns transferable query--reference residuals, and MetaUAS~\citep{gao2024metauas} learns change segmentation from synthetic image pairs using soft feature alignment.
These comparison mechanisms primarily support anomaly scoring and segmentation.
Our work investigates how spatial comparison information can also support autoregressive anomaly reasoning.

\paragraph{LVLM-based anomaly detection and reasoning.}
LVLM-based methods connect anomaly perception with natural-language interpretation.
AnomalyGPT~\citep{gu2024anomalygpt} encodes predicted localization maps as learned prompt embeddings to guide anomaly judgments and dialogue.
MMAD~\citep{jiang2025mmad} provides a benchmark for evaluating diverse anomaly-understanding capabilities.
Recent methods further develop reasoning supervision.
IAD-R1~\citep{li2026iadr1} combines anomaly-specific chain-of-thought training with reinforcement learning, while AD-FM~\citep{liao2026adfm} couples multi-stage inspection with classification and localization rewards.
The MMR-AD dataset~\citep{yao2026mmrad} provides query--reference comparative reasoning annotations; its accompanying Anomaly-R1 model combines supervised fine-tuning with reinforcement learning for detection and localization.
JUDO~\citep{kang2026judo} learns comparative explanations and anomaly segmentation expressed as patch coordinates from juxtaposed query and normal images, together with domain-oriented reasoning.
These methods incorporate visual prompting, comparative explanations, and spatial supervision.
VD-DeepStack explores dense query--reference difference features as an additional source of conditioning for language reasoning.

\paragraph{Visual conditioning for multimodal reasoning.}
Beyond anomaly detection, VC-STaR~\citep{pan2026contrast} uses contrastive visual question-answering pairs to refine training rationales.
Architectural approaches investigate how visual features enter the language model.
Cambrian-1~\citep{tong2024cambrian} aggregates features from multiple vision encoders through spatially structured cross-attention at multiple LLM layers.
DeepStack~\citep{meng2024deepstack} injects additional visual features at multiple decoder depths without extending the input sequence.
Building on this interface, VD-DeepStack supplies dense query--reference difference features alongside visual appearance features at multiple decoder depths, providing explicit comparison information for subsequent autoregressive anomaly reasoning.
\begin{figure*}[tbhp]
\centering
\includegraphics[width=0.99\textwidth]{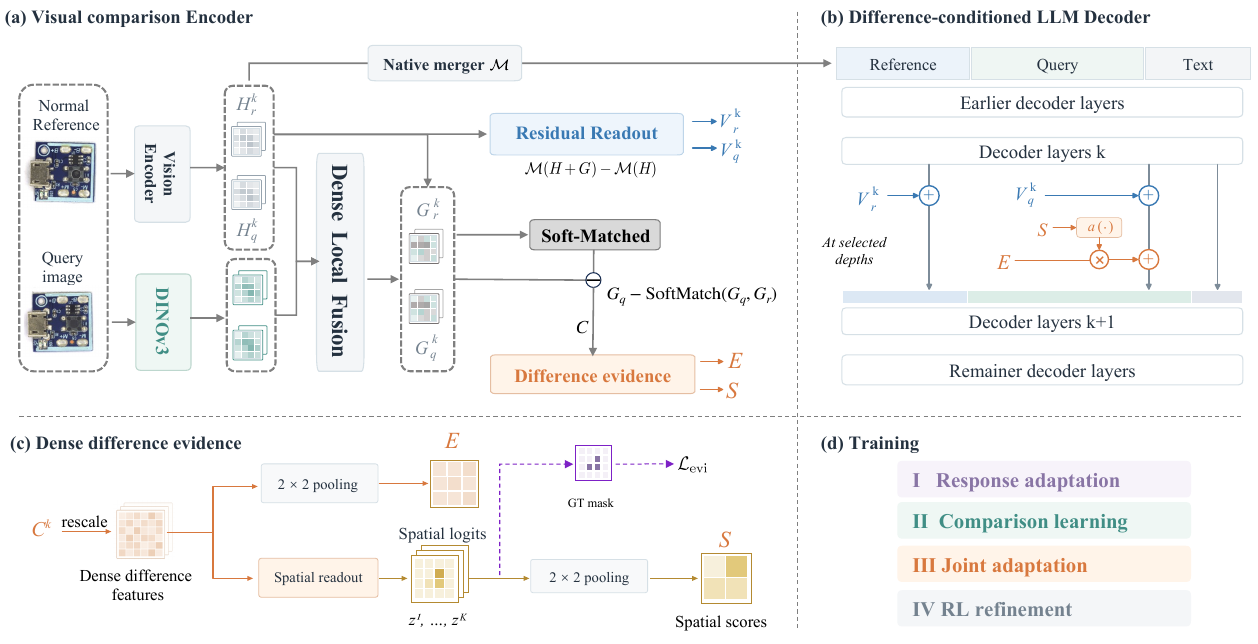}
\caption{\textbf{VD-DeepStack overview.}
Blue paths provide appearance context to both images; orange paths inject difference vectors $E$ only into query tokens, weighted by spatial scores $S$.
Injection occurs during prefill without adding tokens.}
\label{fig:framework}
\end{figure*}
\section{Method}
\label{sec:method}

Visual Difference DeepStack (VD-DeepStack) complements
language-based comparative reasoning with dense query--reference
difference features.
Given a query image $I_q$, a normal reference image $I_r$
from the same object category, and an instruction, the model
generates a reasoning trace followed by an anomaly decision
and, when applicable, defect boxes with type labels.
The method first constructs dense difference features through
visual fusion and reference matching, then injects them into
query-image states at multiple decoder depths, with an auxiliary
path supplying appearance context.
Masks and boxes are used for training supervision, not as
inference inputs.

\subsection{Dense Visual-Difference Evidence}
\label{sec:visual_difference}

\paragraph{Features for fine-grained comparison.}
Query and reference images can share the same object semantics
while differing in local texture, surface appearance, or structure.
To support comparison at this finer level, we supplement the
LVLM's native visual hierarchy with frozen DINOv3
features~\citep{simeoni2025dinov3}.
This choice is informed by studies of visual limitations in
CLIP-based LVLMs and the complementary value of self-supervised
visual representations~\citep{tong2024eyes,tong2024cambrian}.
Appendix~\ref{app:dino_local_features} provides a comparison
of frozen visual encoders under a common anomaly-detection protocol.

Let $H_i^k\in\mathbb R^{N_i\times d_v}$ denote the native
pre-merger visual states of image $i\in\{r,q\}$ at level $k$,
and let $D_i^k$ be a learned mixture of frozen DINO features
on the corresponding spatial grid.
At each patch location $j$, the native feature attends to
a local DINO neighborhood $\mathcal N(j)$:
\begin{equation}
    R_{i,j}^k=
    \operatorname{LocalAttn}_k
    \bigl(H_{i,j}^k,D_{i,\mathcal N(j)}^k\bigr),
    \qquad
    G_{i,j}^k=\frac{\alpha_k}{\kappa_k}R_{i,j}^k.
    \label{eq:visual_residual}
\end{equation}
The fusion adapter at each level is shared by the two images.
Here $\alpha_k$ is a learned coefficient, and $\kappa_k$
is a detached, reference-derived normalization factor shared
across the pair.
The resulting features $G_i^k$ are used for reference comparison
and visual-context construction.
Feature mixing, local attention, and normalization details
are given in Appendix~\ref{app:vd_implementation}.

\paragraph{Query--reference differences.}
Since the query and reference need not be spatially aligned,
we use soft matching to construct a reference counterpart
for each query patch.
Normalized feature maps $\phi_q^k$ and $\phi_r^k$ define
a learned matching space:
\begin{equation}
\begin{aligned}
    s_{jm}^k
      &=\left\langle\phi_q^k(G_{q,j}^k),
                         \phi_r^k(G_{r,m}^k)\right\rangle,
    &p_{jm}^k
      &=\operatorname{softmax}_m(s_{jm}^k/\tau),\\
    \widehat G_{r\to q,j}^k
      &=\sum_m p_{jm}^kG_{r,m}^k,
    &C_j^k
      &=G_{q,j}^k-\widehat G_{r\to q,j}^k.
\end{aligned}
\label{eq:signed_difference}
\end{equation}
Here $\tau$ is a learned temperature.
Matching weights are computed in the learned metric space,
while the reference counterpart is reconstructed from the
original feature vectors.
Thus, $C_j^k$ represents a feature-valued difference at each
query location.

From these differences, we construct spatial evidence features
$E$ for decoder conditioning and a score map $S$ for weighting
their injection.
We first apply a shared channel-wise modulation to the differences:
\begin{equation}
    Z_j^k=C_j^k\odot\left[1+\tanh f_\theta\!\left(
        [\operatorname{LN}(C_j^k);\xi_j^k]\right)\right].
    \label{eq:anchored_modulation}
\end{equation}
The vector $\xi_j^k$ contains maximum matching similarity,
normalized assignment entropy, and assignment peak.
A level-specific readout predicts spatial anomaly logits:
\begin{equation}
    z_j^k=w_k^\top\operatorname{LN}_k(|Z_j^k|)+b_k.
    \label{eq:spatial_readout}
\end{equation}
These logits receive mask supervision during training.

To obtain evidence on the native image-token grid, we average
each non-overlapping $2\times2$ patch cell using $\mathcal P$
and aggregate the $K=4$ levels:
\begin{equation}
    E=\frac1K\sum_k\mathcal P(Z^k),
    \qquad
    S=\frac1K\sum_k\mathcal P(z^k).
    \label{eq:evidence_aggregate}
\end{equation}
At each pooled query location $t$, $E_t$ provides a difference
vector, while $S_t$ supplies a spatial score used to weight
its injection into the decoder.

\subsection{DeepStack-Conditioned Language Reasoning}
\label{sec:deepstack}

To support interpretation of the differences in $E$, we provide
the decoder with these features alongside per-image appearance
context.
The visual-context path supplements the native image-token states
of both reference and query with fine-grained appearance information,
while the difference-evidence path supplies $E$ only to query states.
Both paths use residual updates at multiple decoder depths,
following DeepStack~\citep{meng2024deepstack}.

\paragraph{Visual-context construction.}
For each image and visual level, we use the frozen native
merger $\mathcal M$ to compute the change in its output
induced by $G_i^k$:
\begin{equation}
    V_i^k=\mathcal M(H_i^k+G_i^k)-\mathcal M(H_i^k).
    \label{eq:visual_merger_delta}
\end{equation}
The same merger is used across images and levels.
The resulting $V_i^k$ supplies per-image appearance context,
complementing the query--reference differences in $E$.

\paragraph{Multi-depth residual injection.}
Let $h_{i,t}^{l_k}\in\mathbb R^{d_h}$ denote the hidden state
at image $i$'s pooled position $t$, after decoder block $l_k$
and before injection.
For each path $b\in\{\mathrm{ctx},\mathrm{evi}\}$, a
depth-specific low-rank writer
$\mathcal T_l^b(x)=B_l^bA_l^b\operatorname{LN}_l^b(x)$
maps its input into decoder space.
Spatial weights $a_t$ are obtained by normalizing $S$ within
each query and applying a sigmoid.
The visual-context update is applied first, followed by the
weighted evidence update at query positions:
\begin{equation}
\begin{aligned}
    \widetilde h_{i,t}^{l_k}
      &=h_{i,t}^{l_k}
        +\operatorname{Cap}_{\eta_{\mathrm{ctx}}}\!\left(
          \mathcal T_{l_k}^{\mathrm{ctx}}(V_{i,t}^{k});
          h_{i,t}^{l_k}\right),
        &&i\in\{r,q\},\\
    \widehat h_{q,t}^{l_k}
      &=\widetilde h_{q,t}^{l_k}
        +\operatorname{Cap}_{\eta_{\mathrm{evi}}}\!\left(
          a_t\mathcal T_{l_k}^{\mathrm{evi}}(E_t);
          \widetilde h_{q,t}^{l_k}\right).
\end{aligned}
\label{eq:deep_injection}
\end{equation}
Here $\operatorname{Cap}_{\eta}(u;h)$ limits the update norm
to at most $\eta\|h\|_2$.
Each selected decoder depth receives its corresponding
visual level $V^k$ and the same aggregated evidence $E$,
using independent writers.
The updates modify existing image-token positions during
multimodal prefill, providing context for subsequent
autoregressive generation without adding evidence tokens.
The spatial normalization, gradient handling, and clipping
operations are specified in Appendix~\ref{app:decoder_paths}.

\subsection{Training}
\label{sec:optimization}

Training combines response supervision with spatial supervision
of the difference features.
During \emph{response adaptation}, we first fine-tune the LVLM
on complete target responses.
In \emph{visual comparison pretraining}, we freeze the resulting
LVLM and DINO and train the visual comparison modules using the
unmodulated difference features $C$ and spatial supervision.
The final supervised stage, \emph{joint adaptation}, optimizes
the visual comparison modules, channel-wise modulator, both writer
paths, and decoder LoRA adapters~\citep{hu2022lora} together,
while keeping the base LVLM and DINO weights frozen.

The joint objective is
\begin{equation}
    \mathcal L_U=\mathcal L_{\mathrm{seq}}
      +\underbrace{\mathcal L_{\mathrm{evi}}
        +\lambda_m\mathcal L_{\mathrm{match}}
        +\lambda_o\mathcal L_{\mathrm{orth}}}_{\mathcal L_{\mathrm{vis}}}.
    \label{eq:joint_loss}
\end{equation}
Here $\mathcal L_{\mathrm{seq}}$ is causal cross-entropy over
the assistant response, including the reasoning and final answer.
The evidence loss $\mathcal L_{\mathrm{evi}}$ applies Dice
and focal losses to each level's spatial logits and their
mean before pooling.
The matching loss $\mathcal L_{\mathrm{match}}$ supervises
normal-region correspondences and encourages higher
best-match similarity for normal than anomalous regions.
The term $\mathcal L_{\mathrm{orth}}$ regularizes the shared
DINO-mixing transformations.
These visual losses remain active during joint adaptation.
Complete loss definitions and optimization details are
provided in Appendix~\ref{app:vd_implementation}.

In the  \emph{RL refinement} stage, we apply GRPO to the jointly adapted model, updating only
the language backbone.
The reward combines decision correctness, structural validity,
and a localization bonus.
Data selection, reward definitions, and training settings are
provided in Appendix~\ref{app:rl_refinement}.

\section{Experiments}
\subsection{Implementation Details}
\label{sec:implementation_details}

Experiments use eight NVIDIA H200 GPUs, with Qwen2.5-VL-7B or Qwen3-VL-8B   as the LVLM
and frozen DINOv3 ViT-L/16 as the complementary encoder.
The three supervised stages use $10\%$, $25\%$, and $100\%$ of the
training set, respectively. Stage III oversamples normal examples
threefold to mitigate class imbalance. Each stage runs for one epoch
over its resulting training list. Stage I fine-tunes the complete LVLM at $10^{-5}$. Stage II freezes
the LVLM and trains the visual comparison modules with module-specific peak
learning rates of $2\times10^{-5}$--$10^{-4}$. Stage III keeps the
LVLM backbone frozen and jointly updates the visual comparison modules
($2\times10^{-6}$--$10^{-5}$), the channel-wise modulator and both writer
paths ($2\times10^{-5}$), and decoder LoRA adapters ($10^{-5}$).
Detailed parameter groups, interface settings, and numerical constants
are provided in Appendix~\ref{app:vd_implementation}.

\begin{table}[h]
\centering
\caption{
Detection/localization (\%) on four benchmarks.
Bold and underline mark the best and second-best values, respectively,
for detection and localization separately, including VD-DeepStack
and excluding full-shot reference methods. Ties share the same rank.
VD-DeepStack results include RL refinement.
IAD-R1$^*$ is our reproduction trained and evaluated using MMR-AD data.
Dashes denote unavailable results.
}
\label{tab:main_results}

\setlength{\tabcolsep}{2.7pt}
\renewcommand{\arraystretch}{1.14}

\resizebox{\textwidth}{!}{%
\begin{tabular}{@{}l*{12}{c}@{}}
\toprule
\rowcolor{vdTableHead}
\multirow{2}{*}{\textbf{Model}}
& \multicolumn{3}{c}{\textbf{MVTec AD}}
& \multicolumn{3}{c}{\textbf{VisA}}
& \multicolumn{3}{c}{\textbf{MVTec 3D}}
& \multicolumn{3}{c}{\textbf{MPDD}} \\
\cmidrule(lr){2-4}
\cmidrule(lr){5-7}
\cmidrule(lr){8-10}
\cmidrule(lr){11-13}
\rowcolor{vdTableHead}
& Acc. & Recall & Prec.
& Acc. & Recall & Prec.
& Acc. & Recall & Prec.
& Acc. & Recall & Prec. \\
\midrule

\rowcolor{vdTableGroup}
\multicolumn{13}{@{}l@{}}{\textbf{Commercial MLLMs}} \\
Gemini-2.5-pro
& 79.4/34.4
& \underline{98.4}/46.0
& 79.0/48.8
& 65.7/10.5
& \underline{97.2}/17.1
& 62.7/21.3
& 75.7/10.9
& \underline{90.1}/18.3
& 83.2/19.9
& 62.1/16.3
& 92.2/23.2
& 60.8/26.7 \\

GPT-4o
& 68.9/8.1
& 74.1/12.9
& 82.4/16.9
& 57.6/3.7
& 69.4/5.8
& 61.1/9.2
& 67.8/7.0
& 74.0/11.5
& 83.4/14.8
& 63.1/13.6
& 84.3/20.4
& 64.3/22.3 \\

GPT-5
& 78.7/41.8
& 95.9/64.2
& 79.4/53.0
& 65.8/18.5
& 94.5/31.0
& 63.3/30.7
& 75.0/27.2
& 85.3/41.1
& 84.1/43.2
& 68.4/21.2
& 91.9/34.5
& 67.0/29.2 \\

Qwen3.8-Flash
& 76.0/58.0
& 95.1/71.5
& 72.2/73.4
& 78.9/32.0
& 86.3/43.2
& 76.2/53.0
& \textbf{78.3}/29.9
& 82.2/40.1
& 90.4/46.7
& 63.0/37.9
& 79.9/49.7
& 62.2/52.5 \\

ChatGPT-5.6-Luna
& 86.0/57.9
& 91.9/79.1
& 84.6/67.6
& \textbf{83.6}/44.3
& 76.0/\underline{56.5}
& \textbf{91.0}/68.0
& 69.5/37.4
& 68.8/55.6
& 95.0/57.2
& 73.4/39.7
& 67.6/\textbf{60.5}
& \textbf{79.3}/52.0 \\

\addlinespace[3pt]
\rowcolor{vdTableGroup}
\multicolumn{13}{@{}l@{}}{\textbf{Full-shot anomaly detectors (reference)}} \\

PaDiM
& 93.4/65.0
& 96.2/84.7
& 94.8/74.0
& 81.5/34.1
& 80.6/48.6
& 85.2/51.9
& 82.2/32.4
& 97.0/44.8
& 84.1/52.8
& 74.7/28.1
& 96.4/48.2
& 72.6/37.4 \\

PatchCore
& 93.8/73.2
& 97.5/82.7
& 94.6/86.8
& 83.0/44.9
& 80.3/51.2
& 90.7/78.0
& 81.4/39.7
& 96.6/46.2
& 83.5/69.8
& 86.3/57.3
& 84.1/62.9
& 92.3/76.5 \\

HGAD
& 96.2/73.2
& 96.1/86.3
& 98.3/82.4
& 90.6/52.7
& 91.5/64.0
& 92.1/75.5
& 84.5/45.5
& 95.4/56.6
& 87.6/64.8
& 86.5/34.2
& 81.8/51.6
& 90.2/44.2 \\

Dinomaly
& 94.1/58.5
& 93.2/84.3
& 97.8/66.0
& 79.8/34.3
& 69.8/47.5
& 96.0/54.2
& 82.0/44.3
& 93.1/62.8
& 86.7/57.1
& 87.5/56.0
& 88.3/64.7
& 91.0/69.0 \\

INP-Former
& 95.7/59.2
& 97.5/84.6
& 96.9/66.2
& 83.2/36.2
& 78.1/51.2
& 93.4/53.7
& 84.8/48.2
& 93.2/58.0
& 89.3/69.0
& 87.9/49.4
& 85.1/65.9
& 94.2/58.4 \\

\addlinespace[3pt]
\rowcolor{vdTableGroup}
\multicolumn{13}{@{}l@{}}{\textbf{Open-source MLLMs}} \\

Llama4-Maverick
& 76.1/19.9
& 83.2/28.4
& 83.5/36.2
& 63.8/6.7
& 71.6/10.3
& 66.2/15.4
& 67.1/6.1
& 69.1/9.6
& 86.7/13.9
& 64.1/10.0
& 70.7/12.7
& 68.8/18.0 \\

Gemma3-27B
& 74.4/4.0
& \textbf{99.6}/8.5
& 74.2/7.0
& 58.5/0.5
& \textbf{98.6}/1.4
& 57.3/0.9
& 75.1/0.7
& \textbf{98.4}/1.6
& 75.7/1.3
& 59.0/3.5
& \textbf{99.1}/7.5
& 58.7/5.2 \\

Qwen2.5-VL-7B
& 75.0/8.9
& 70.8/12.5
& 92.4/20.8
& 65.9/2.2
& 50.4/3.0
& 77.1/8.7
& 59.9/1.2
& 55.1/1.8
& 90.9/3.6
& 63.6/7.2
& 61.2/9.1
& 71.7/15.1 \\

Qwen2.5-VL-72B
& 83.9/36.4
& 94.4/47.8
& 85.6/53.0
& 70.0/11.1
& 79.1/15.9
& 71.0/25.5
& 71.6/13.3
& 75.8/19.2
& 86.4/27.0
& 67.0/24.7
& 77.0/32.4
& 71.3/37.9 \\

Qwen3-VL-8B
& 77.1/30.0
& 82.4/58.3
& 77.8/38.5
& 77.2/15.1
& 72.6/26.5
& 82.6/24.7
& 64.0/14.3
& 60.0/28.7
& 93.6/21.8
& 67.4/20.6
& 75.3/42.9
& 66.5/25.1 \\

Qwen3-VL-32B
& 83.9/64.5
& 92.9/\underline{80.6}
& 81.5/74.8
& 81.0/31.0
& 79.8/40.4
& 83.7/52.5
& 73.6/43.3
& 74.7/53.4
& 91.2/64.9
& 69.8/44.5
& 76.9/\underline{59.5}
& 71.4/55.8 \\

Qwen3.5-9B
& 70.9/28.1
& 71.3/43.3
& 73.9/42.7
& 65.4/9.4
& 71.4/16.1
& 64.9/18.9
& 65.5/8.5
& 66.5/14.0
& 87.9/15.3
& 59.6/15.5
& 55.7/26.0
& 64.0/25.6 \\

Qwen3.8-27B
& 78.5/\textbf{74.7}
& 96.5/\textbf{82.1}
& 73.3/\underline{87.5}
& 80.3/43.1
& 89.9/54.2
& 77.8/66.6
& 75.7/45.5
& 85.6/55.7
& 85.1/64.7
& 66.4/40.0
& \underline{92.7}/53.9
& 61.3/50.5 \\

InternVL3.5-8B
& 76.5/5.4
& 81.3/9.7
& 85.4/10.6
& 67.8/2.2
& 66.1/3.3
& 72.4/5.5
& 65.8/1.1
& 66.7/1.9
& 86.9/2.4
& 64.3/4.1
& 64.5/5.8
& 69.8/8.7 \\

InternVL3.5-38B
& 84.9/7.1
& 95.6/13.2
& 85.8/12.4
& 74.1/1.8
& 87.5/3.2
& 72.4/3.8
& 74.4/1.4
& 83.8/2.6
& 85.6/2.9
& 70.2/5.1
& 87.5/10.1
& 71.0/8.3 \\

\addlinespace[3pt]
\rowcolor{vdTableGroup}
\multicolumn{13}{@{}l@{}}{\textbf{Task-specific anomaly models}} \\
AnomalyGPT-7B
& 85.7/--
& 86.5/--
& 90.6/--
& 72.8/--
& 67.6/--
& 79.8/--
& 64.7/--
& 68.6/--
& 84.6/--
& 67.6/--
& 73.2/--
& 58.9/-- \\

Anomaly-R1-7B
& \underline{88.8}/66.3
& 91.9/74.2
& 92.8/84.2
& 74.9/36.4
& 72.3/41.2
& 82.7/77.3
& 74.6/48.5
& 73.8/\underline{56.0}
& 92.8/78.8
& 70.8/35.9
& 74.2/45.6
& \underline{77.6}/52.8 \\

IAD-R1$^*$
& 83.0/56.7
& 92.0/63.2
& 79.9/80.2
& 78.5/40.4
& 67.7/44.1
& \underline{88.7}/81.4
& 72.8/42.8
& 73.9/50.0
& 91.6/75.0
& 72.9/35.9
& 71.4/38.9
& 75.2/63.6 \\

AD-FM-7B
& --/--
& --/--
& --/--
& 77.4/--
& --/--
& --/--
& --/--
& --/--
& --/--
& 72.7/--
& --/--
& --/-- \\
\addlinespace[3pt]
\midrule
\rowcolor{vdTableOursRow}
VD-DeepStack-7B
& \textbf{91.0}/66.6
& 93.9/74.0
& \textbf{93.2}/85.0
& 82.1/\underline{47.9}
& 79.6/52.4
& 86.5/\underline{83.3}
& 76.3/\underline{51.9}
& 73.9/55.1
& \textbf{98.3}/\underline{79.9}
& \textbf{78.6}/\textbf{54.8}
& 81.1/58.2
& \textbf{79.3}/\textbf{76.4} \\
\rowcolor{vdTableOursRow}

VD-DeepStack-8B
& \textbf{91.0}/\underline{70.4}
& 93.2/75.8
& \underline{93.0}/\textbf{89.7}
& \underline{82.7}/\textbf{54.9}
& 81.3/\textbf{57.6}
& 87.5/\textbf{90.7}
& \underline{76.6}/\textbf{55.7}
& 74.2/\textbf{57.8}
& \underline{98.1}/\textbf{84.0}
& \underline{75.2}/\underline{49.8}
& 72.5/54.8
& \textbf{79.3}/\underline{72.1} \\

\bottomrule
\end{tabular}%
}
\end{table}

\subsection{Datasets and Evaluation Metrics}

\paragraph{Datasets.}
We use the same industrial datasets as MMR-AD~\citep{yao2026mmrad},
comprising 14 datasets: MVTec AD~\citep{DBLP:conf/cvpr/BergmannFSS19},
MVTec 3D-AD~\citep{DBLP:conf/visapp/BergmannJSS22},
VisA~\citep{DBLP:conf/eccv/ZouJPZD22},
MPDD~\citep{DBLP:conf/icumt/JezekJBDS21},
MVTec LOCO~\citep{DBLP:journals/ijcv/BergmannBFSS22},
GoodsAD~\citep{DBLP:journals/ral/ZhangDBD24},
Real-IAD~\citep{DBLP:conf/cvpr/WangZGGZGQCM24},
Real-IAD D$^3$~\citep{DBLP:conf/cvpr/ZhuWZWPZCCGZGWC25},
MANTA~\citep{DBLP:conf/cvpr/0007FHDDYP025},
MIAD~\citep{DBLP:conf/iccvw/BaoCLWFWZZ23},
CableInspect-AD~\citep{DBLP:conf/nips/ArodiLBZLPBM24},
WFDD~\citep{DBLP:conf/eccv/ChenLLZ24},
Texture-AD~\citep{DBLP:journals/corr/abs-2409-06367},
and 3CAD~\citep{DBLP:conf/aaai/YangXS0ML025}.
We evaluate on four industrial benchmarks: MVTec AD, MVTec 3D-AD,
VisA, and MPDD. We evaluate VD-DeepStack under a 1-shot reference setting:
each query image is paired with one normal reference image
from the same object category. For each industrial test benchmark, the entire benchmark is excluded from training.

To assess cross-domain generalization from industrial inspection to
medical imaging, we additionally evaluate on
HeadCT and BrainMRI, which have been used as medical anomaly detection
benchmarks~\citep{DBLP:conf/cvpr/SalehiSBRR21}.
Both datasets provide image-level labels but lack pixel-level anomaly
masks, and are therefore used only for image-level anomaly detection. HeadCT and BrainMRI are also evaluated under a 1-shot reference
setting. 

\paragraph{Evaluation metrics.}
Following MMR-AD~\citep{yao2026mmrad},
we report accuracy, recall, and precision for both image-level anomaly
detection and bounding-box localization. Image-level predictions are
obtained from the model-generated normal/abnormal decisions, with
anomalous images treated as the positive class. For localization, we follow MMR-AD and count a predicted bounding box
as correct when its intersection over union (IoU) with a ground-truth
box is at least 0.1. This relatively permissive threshold measures
coarse anomaly localization rather than precise box alignment. For each industrial
benchmark, we average results across its categories. HeadCT and BrainMRI
are evaluated only using the image-level detection metrics.

\subsection{Main Results}
\label{sec:main_results}

\paragraph{Compared methods.}
Table~\ref{tab:main_results} compares VD-DeepStack with commercial
LVLMs, including Gemini and GPT models, and open-source LVLMs from
the Llama, Gemma, InternVL, and Qwen families.
We also include task-specific anomaly models:
AnomalyGPT~\citep{gu2024anomalygpt},
Anomaly-R1~\citep{yao2026mmrad},
IAD-R1~\citep{li2026iadr1}, and AD-FM~\citep{liao2026adfm}.
IAD-R1$^*$ denotes our reproduction of IAD-R1, retrained and
evaluated using MMR-AD data.
Anomaly-R1 is particularly relevant to our comparison because it
learns anomaly reasoning from query--reference comparative annotations.
PaDiM, PatchCore, HGAD, Dinomaly, and INP-Former are additionally
reported as full-shot reference methods under a different training
setting.
\begin{table}[h]
\centering
\small
\caption{
Ablation of visual-difference conditioning.
Entries report detection/localization (\%) at localization IoU$\geq$0.1.
Ctx and Evi denote the visual-context and difference-evidence paths.
All variants except VD-DeepStack-7B are trained without RL.
Bold marks the best result among the variants without RL.
}
\label{tab:evidence_ablation}
\begin{tabular}{@{}l*{6}{c}@{}}
\toprule
\multirow{2}{*}{\textbf{Variant}}
& \multicolumn{3}{c}{\textbf{VisA}}
& \multicolumn{3}{c}{\textbf{MVTec 3D}} \\
\cmidrule(lr){2-4}\cmidrule(lr){5-7}
& Acc. & Rec. & Prec. & Acc. & Rec. & Prec. \\
\midrule
Baseline
& 78.0/38.3 & 66.5/42.0 & 91.5/81.6
& 70.4/38.7 & 67.7/45.4 & 93.0/72.5 \\

w/o Evi
& 78.9/43.4 & 69.6/46.9 & 90.4/84.0
& 67.7/43.2 & 64.7/47.2 & \textbf{97.3}/75.1 \\
E-tokens
& 78.3/41.0 & 67.3/44.7 & 91.7/82.2
& 71.9/44.8 & 69.0/48.6 & 94.9/75.7 \\
S-only
& 79.7/44.3 & 69.3/47.5 & \textbf{92.8}/\textbf{86.3}
& 72.4/47.0 & 68.7/50.6 & 95.9/78.6 \\
w/o Ctx
& \textbf{81.8}/45.2 & 76.3/49.6 & 91.1/81.7
& 74.2/48.6 & 70.7/52.6 & 96.2/78.9 \\

w/o Ctx \& DINO 
& 79.6/40.8 & 67.6/44.4 & 90.1/81.5
& 70.3/43.2 & 67.9/47.5 & 97.1/74.6 \\

VD-DeepStack-7B w/o RL
& 81.4/\textbf{46.4} & \textbf{76.6}/\textbf{50.6} & 90.4/82.4
& \textbf{75.8}/\textbf{50.0} & \textbf{72.4}/\textbf{53.8} & 96.6/\textbf{79.5} \\
\midrule
VD-DeepStack-7B
& 82.1/47.9 & 79.6/52.4 & 86.5/83.3
& 76.3/51.9 & 73.9/55.1 & 98.3/79.9 \\
\bottomrule
\end{tabular}
\end{table}
\paragraph{Detection and localization performance.}
VD-DeepStack-7B improves both detection and localization accuracy
over Anomaly-R1-7B on all four benchmarks.
Averaged equally across the four benchmarks, it achieves
82.0\% detection accuracy and 55.3\% localization accuracy,
exceeding Anomaly-R1-7B by 4.7 and 8.5 percentage points, respectively.
The gains are particularly pronounced on VisA and MPDD:
detection accuracy increases by 7.2 and 7.8 points,
while localization accuracy improves by 11.5 and 18.9 points.
On both datasets, localization recall and precision also improve,
showing that the gains extend to both measures of region prediction.
The 7B variant also exceeds IAD-R1$^*$ on both accuracy measures across all
four benchmarks, with average gains of 5.2 and 11.4 points.
Relative to the general-purpose LVLMs in the table, the 7B variant
achieves higher localization accuracy on VisA, MVTec 3D, and MPDD,
although individual general-purpose models remain stronger on
some detection metrics.
Under the adopted IoU threshold, these results demonstrate
improvements in identifying anomalous images and coarsely
localizing their defects.

\paragraph{Results across LVLM backbones.}
Both the Qwen2.5-VL-7B and Qwen3-VL-8B instantiations improve
detection and localization accuracy over their respective base
LVLMs on every benchmark.
The 8B variant achieves localization accuracies of 54.9\% on VisA
and 55.7\% on MVTec 3D, the highest reported values in the table
for these two benchmarks.
The 7B variant remains stronger on MPDD, indicating that the
relative performance of the two instantiations depends on the dataset.
Together, these results support the effectiveness of VD-DeepStack
across both backbones and are consistent with our motivation to
make visual comparison explicitly available to language reasoning.
The contributions of the individual components are examined
in Table~\ref{tab:evidence_ablation}.
\subsection{Ablation Studies}
\label{sec:ablation}

Table~\ref{tab:evidence_ablation} compares visual-conditioning variants
without RL and reports the complete model after RL refinement.
\emph{Baseline} removes both paths; \emph{w/o Evi} removes evidence
injection, while \emph{w/o Ctx} removes context injection but retains
the upstream comparison features.
\emph{w/o Ctx \& DINO} additionally removes DINO and constructs
difference evidence from native Qwen visual features.
With context retained, \emph{S-only} feeds only $S$ to the evidence
writers, whereas \emph{E-tokens} encodes $E$ and $S$ as 16 context
tokens in place of deep evidence injection.
\emph{VD-DeepStack-7B w/o RL} retains both paths after joint adaptation;
\emph{VD-DeepStack-7B} additionally includes RL refinement.

\emph{VD-DeepStack-7B w/o RL} improves detection/localization accuracy over
\emph{w/o Evi}
by 2.5/3.0 points on VisA and 8.1/6.8 on MVTec 3D, and also exceeds
\emph{S-only} on both datasets.
Its localization accuracy surpasses \emph{E-tokens} by 5.4 and 5.2
points, although this comparison changes both spatial compression
and the injection interface.
\begin{table}[h]
\small
\centering
\caption{Cross-domain anomaly detection performance on medical datasets (\%). Results are means over five 1-shot runs with different normal reference images; standard deviations are shown where available. Average denotes the unweighted mean across the two datasets.}
\label{tab:medical_cross_domain}

\setlength{\tabcolsep}{3pt}
\renewcommand{\arraystretch}{1.08}

\resizebox{\linewidth}{!}{%
\begin{tabular}{@{}lccccccccc@{}}
\toprule
\multirow{2}{*}{\textbf{Model}}
& \multicolumn{3}{c}{\textbf{BrainMRI}}
& \multicolumn{3}{c}{\textbf{HeadCT}}
& \multicolumn{3}{c}{\textbf{Average}} \\
\cmidrule(lr){2-4}
\cmidrule(lr){5-7}
\cmidrule(lr){8-10}
& Acc. & Recall & Prec.
& Acc. & Recall & Prec.
& Acc. & Recall & Prec. \\
\midrule

Qwen3.8-Flash
& 71.4$\pm$4.7 & 88.1$\pm$2.6 & 71.9$\pm$4.2
& 50.5$\pm$2.2 & 86.4$\pm$2.1 & 50.3$\pm$1.3
& 60.9 & 87.3 & 61.1 \\

Qwen3.8-27B
& 50.4$\pm$7.3 & 39.2$\pm$10.5 & 65.8$\pm$9.1
& 53.6$\pm$2.2 & 13.8$\pm$4.7 & 68.1$\pm$8.6
& 52.0 & 26.5 & 67.0 \\

IAD-R1$^*$
& 78.7$\pm$5.6 & 92.4$\pm$1.0 & 77.6$\pm$5.4
& 73.5$\pm$4.0 & 97.0$\pm$2.5 & 66.2$\pm$3.6
& 76.1 & 94.7 & 71.9 \\

\midrule
VD-DeepStack-7B
& 83.9$\pm$2.4 & 94.3$\pm$0.8 & 82.2$\pm$2.9
& 77.1$\pm$2.0 & 95.8$\pm$1.6 & 69.8$\pm$2.1
& 80.5 & 95.1 & 76.0 \\

VD-DeepStack-8B
& 87.7$\pm$1.2 & 98.5$\pm$0.7 & 84.2$\pm$1.6
& 77.0$\pm$3.8 & 99.8$\pm$0.4 & 68.8$\pm$3.7
& 82.3 & 99.1 & 76.5 \\

\bottomrule
\end{tabular}%
}
\end{table}
\begin{table}[h]

\centering
\small
\caption{
Quality of generated CoT reasoning scored by GLM-5.3-Flash
(0--100; higher is better).
The judge compares CoT text with ground-truth annotations,
excluding final answers.
Abn./Norm.\ denote abnormal/normal samples.
Overall is the mean score over all evaluated samples.
Bold indicates the best score in each column.
}
\label{tab:llm_judge_scores}

\begin{tabular}{@{}l*{9}{c}@{}}
\toprule
\multirow{2}{*}{\textbf{Model}}
& \multicolumn{2}{c}{\textbf{MVTec AD}}
& \multicolumn{2}{c}{\textbf{MVTec 3D}}
& \multicolumn{2}{c}{\textbf{VisA}}
& \multicolumn{2}{c}{\textbf{MPDD}}
& \multirow{2}{*}{\textbf{Overall}} \\
\cmidrule(lr){2-3}
\cmidrule(lr){4-5}
\cmidrule(lr){6-7}
\cmidrule(lr){8-9}
& Abn. & Norm.
& Abn. & Norm.
& Abn. & Norm.
& Abn. & Norm.
& \\
\midrule

IAD-R1$^*$
& 69.9 & 75.9
& 51.2 & 81.9
& 49.8 & \textbf{90.1}
& 58.2 & 73.9
& 68.3 \\

Qwen3.8-Flash
& \textbf{74.7} & 57.5
& \textbf{58.1} & 61.4
& \textbf{62.4} & 72.2
& \textbf{66.5} & 57.7
& 65.6 \\

VD-DeepStack-7B
& 68.1 & \textbf{86.1}
& 55.6 & \textbf{86.6}
& 56.2 & 87.8
& 60.0 & \textbf{74.4}
& \textbf{69.4} \\

\bottomrule
\end{tabular}
\end{table}

With context injection disabled, DINO-enhanced comparison features
improve detection/localization accuracy over native Qwen features
by 2.2/4.4 points on VisA and 3.9/5.4 on MVTec 3D.
Visual context adds 1.2 and 1.4 points in localization accuracy,
with mixed detection effects.
These results suggest that feature-valued differences provide
useful information beyond spatial scores alone. Removing evidence
injection produces larger localization drops than removing context
on both datasets, supporting the importance of the difference path
while showing that appearance context can further improve
localization when difference evidence is retained.
Relative to \emph{VD-DeepStack-7B w/o RL}, \emph{VD-DeepStack-7B} improves
detection/localization accuracy on both datasets,
with a detection recall--precision trade-off on VisA.
\subsection{Generalization and Reasoning Quality}
\label{sec:additional_evaluation}

\paragraph{Cross-domain detection.}
VD-DeepStack's gains extend from industrial inspection to medical
imaging under 1-shot evaluation (Table~\ref{tab:medical_cross_domain}).
The 7B and 8B variants achieve dataset-averaged accuracies of
80.49\% and 82.3\%, respectively, exceeding IAD-R1$^*$ (76.08\%),
Qwen3.8-Flash (60.94\%), and Qwen3.8-27B (51.98\%).
Both variants improve accuracy and precision over IAD-R1$^*$ on
each dataset, while the 8B variant achieves an average recall of
99.1\% with an average precision of 76.5\%.

\paragraph{Quality of CoT reasoning.}
To assess CoT quality under visual-difference conditioning, we use
GLM-5.3-Flash to evaluate the semantic consistency of generated CoTs
with ground-truth annotations, excluding final answers.
The scoring protocol is provided in Appendix~\ref{app:llm_judge_protocol}.
In Table~\ref{tab:llm_judge_scores}, VD-DeepStack-7B achieves the
highest overall score (69.4), exceeding IAD-R1$^*$ on six of eight
subsets, while Qwen3.8-Flash scores higher on abnormal samples.

\begin{figure*}[h]
\centering
\includegraphics[width=0.97\textwidth]{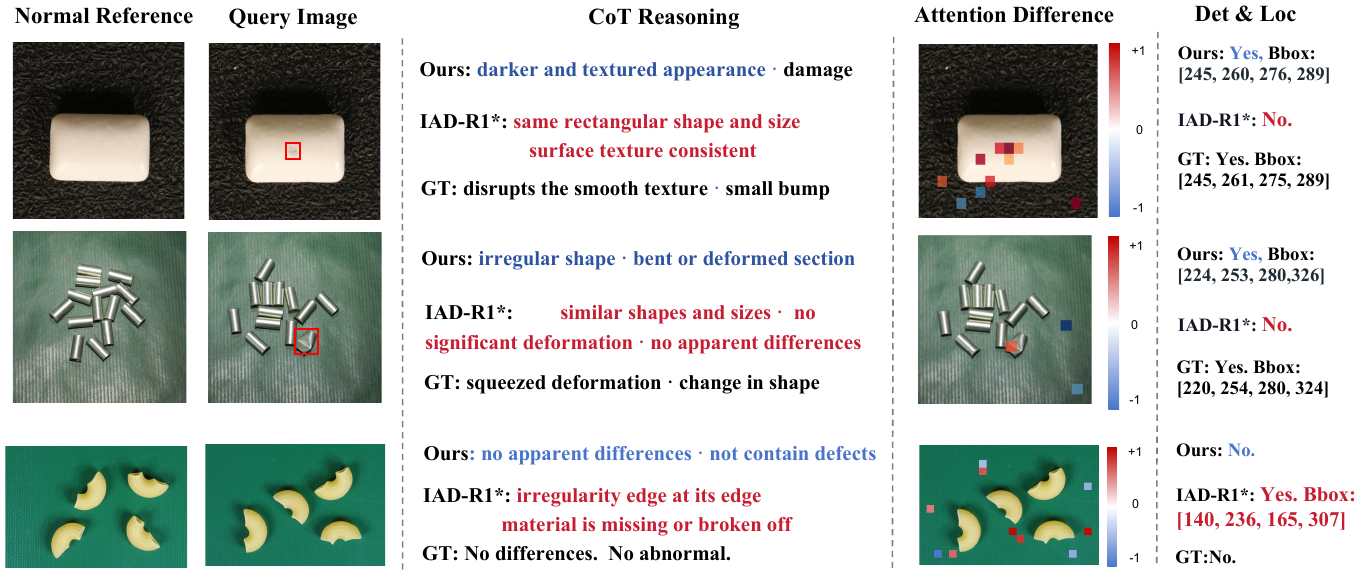}
\caption{Qualitative comparison with IAD-R1$^*$. Attention is extracted from the token position immediately preceding the final answer, after reasoning, to query-image tokens and averaged over the last four LLM layers. Attention differences are computed as VD-DeepStack minus IAD-R1$^*$; red and blue indicate increased and decreased attention, respectively.}
\label{fig:qualitative}
\end{figure*}

\subsection{Qualitative Analysis}
Figure~\ref{fig:qualitative} connects visual attention with
generated descriptions and predictions.
In the two anomalous examples, attention increases relative to
IAD-R1$^*$ overlap the texture defect and crushed component.
These changes at the pre-answer position align with
VD-DeepStack's descriptions of altered texture and deformation,
and its predicted boxes closely match GT.
IAD-R1$^*$ instead emphasizes overall shape similarity and
misses both defects.
On normal pasta, the baseline mistakes an intact edge for
missing material, whereas VD-DeepStack correctly reports no
defect; the attention differences are more spatially dispersed.  On normal pasta, the baseline mistakes an intact edge for
missing material, whereas VD-DeepStack correctly reports no defect; the attention differences are
more spatially dispersed.
In these abnormal examples, VD-DeepStack attends more strongly
to the defect regions than IAD-R1$^*$.

\section{Conclusion}
Motivated by a potential visual comparison--reasoning gap, we
introduced VD-DeepStack for few-shot anomaly detection.
The framework constructs dense query--reference difference
evidence and injects it into existing query-image states through
residual updates at multiple decoder depths, without adding
evidence tokens. An auxiliary visual-context path supplies
appearance features to both reference and query states.
Experiments with two LVLM backbones demonstrate detection and
localization gains across four industrial benchmarks, with
additional detection gains on two medical datasets supporting
cross-domain generalization. Ablations support the value of
difference vectors beyond spatial scores alone, while CoT
evaluation indicates improved overall semantic agreement with
ground-truth annotations. Together, these findings suggest that
connecting fine-grained visual comparison with language reasoning
benefits from the joint design of difference representations
and their integration into the decoder.

\section*{AI Use Statement}
The authors used ChatGPT (OpenAI) for language polishing and suggestions on figure layout and presentation. The research methodology, experiments, analysis, and conclusions were independently developed and verified by the authors.

\bibliography{iclr2027_conference}

@article{li2026iadr1,
  title   = {{IAD-R1}: Reinforcing Consistent Reasoning in Industrial Anomaly Detection},
  author  = {Li, Yanhui and Cao, Yunkang and Liu, Chengliang and Xiong, Yuan and Dong, Xinghui and Huang, Chao},
  journal = {Proceedings of the AAAI Conference on Artificial Intelligence},
  volume  = {40},
  number  = {8},
  pages   = {6583--6591},
  year    = {2026},
  doi     = {10.1609/aaai.v40i8.37588}
}

@article{liao2026adfm,
  title   = {{AD-FM}: Multimodal {LLM}s for Anomaly Detection via Multi-Stage Reasoning and Fine-Grained Reward Optimization},
  author  = {Liao, Jingyi and Su, Yongyi and Tu, Rong-Cheng and Jin, Zhao and Sun, Wenhao and Li, Yiting and Xu, Xun and Tao, Dacheng and Yang, Xulei},
  journal = {Proceedings of the AAAI Conference on Artificial Intelligence},
  volume  = {40},
  number  = {18},
  pages   = {15234--15242},
  year    = {2026},
  doi     = {10.1609/aaai.v40i18.38548}
}

@inproceedings{yao2026mmrad,
  title     = {{MMR-AD}: A Large-Scale Multimodal Dataset for Benchmarking General Anomaly Detection with Multimodal Large Language Models},
  author    = {Yao, Xincheng and Qian, Zefeng and Shi, Chao and Song, Jiayang and Zhang, Chongyang},
  booktitle = {Proceedings of the IEEE/CVF Conference on Computer Vision and Pattern Recognition},
  pages     = {43072--43082},
  month     = {June},
  year      = {2026}
}

@inproceedings{kang2026judo,
  title     = {{JUDO}: A Juxtaposed Domain-Oriented Multimodal Reasoner for Industrial Anomaly {QA}},
  author    = {Kang, Hyunju and Lee, Woohyun and Kim, Jaewon and Park, Hogun},
  booktitle = {The Fourteenth International Conference on Learning Representations},
  year      = {2026}
}

@inproceedings{pan2026contrast,
  title     = {Through the Lens of Contrast: Self-Improving Visual Reasoning in {VLM}s},
  author    = {Pan, Zhiyu and Wu, Yizheng and Hua, Jiashen and Feng, Junyi and Yan, Shaotian and Deng, Bing and Cao, Zhiguo and Ye, Jieping},
  booktitle = {The Fourteenth International Conference on Learning Representations},
  year      = {2026}
}

@inproceedings{huang2022regad,
  title     = {Registration Based Few-Shot Anomaly Detection},
  author    = {Huang, Chaoqin and Guan, Haoyan and Jiang, Aofan and Zhang, Ya and Spratling, Michael and Wang, Yanfeng},
  booktitle = {European Conference on Computer Vision},
  year      = {2022},
  doi       = {10.1007/978-3-031-20053-3_18}
}

@inproceedings{fang2023fastrecon,
  title     = {FastRecon: Few-Shot Industrial Anomaly Detection via Fast Feature Reconstruction},
  author    = {Fang, Zheng and Wang, Xiaoyang and Li, Haocheng and Liu, Jiejie and Hu, Qiugui and Xiao, Jimin},
  booktitle = {Proceedings of the IEEE/CVF International Conference on Computer Vision},
  pages     = {17481--17490},
  year      = {2023},
  doi       = {10.1109/ICCV51070.2023.01603}
}

@inproceedings{jeong2023winclip,
  title     = {WinCLIP: Zero-/Few-Shot Anomaly Classification and Segmentation},
  author    = {Jeong, Jongheon and Zou, Yang and Kim, Taewan and Zhang, Dongqing and Ravichandran, Avinash and Dabeer, Onkar},
  booktitle = {Proceedings of the IEEE/CVF Conference on Computer Vision and Pattern Recognition},
  pages     = {19606--19616},
  year      = {2023},
  doi       = {10.1109/CVPR52729.2023.01878}
}

@inproceedings{li2024promptad,
  title     = {PromptAD: Learning Prompts with only Normal Samples for Few-Shot Anomaly Detection},
  author    = {Li, Xiaofan and Zhang, Zhizhong and Tan, Xin and Chen, Chengwei and Qu, Yanyun and Xie, Yuan and Ma, Lizhuang},
  booktitle = {Proceedings of the IEEE/CVF Conference on Computer Vision and Pattern Recognition},
  pages     = {16838--16848},
  year      = {2024}
}

@inproceedings{zhou2024anomalyclip,
  title     = {AnomalyCLIP: Object-agnostic Prompt Learning for Zero-shot Anomaly Detection},
  author    = {Zhou, Qihang and Pang, Guansong and Tian, Yu and He, Shibo and Chen, Jiming},
  booktitle = {International Conference on Learning Representations},
  year      = {2024}
}

@inproceedings{cao2024adaclip,
  title     = {AdaCLIP: Adapting CLIP with Hybrid Learnable Prompts for Zero-Shot Anomaly Detection},
  author    = {Cao, Yunkang and Zhang, Jiangning and Frittoli, Luca and Cheng, Yuqi and Shen, Weiming and Boracchi, Giacomo},
  booktitle = {European Conference on Computer Vision},
  year      = {2024},
  doi       = {10.1007/978-3-031-72761-0_4}
}

@inproceedings{gu2024filo,
  title     = {FiLo: Zero-Shot Anomaly Detection by Fine-Grained Description and High-Quality Localization},
  author    = {Gu, Zhaopeng and Zhu, Bingke and Zhu, Guibo and Chen, Yingying and Li, Hao and Tang, Ming and Wang, Jinqiao},
  booktitle = {Proceedings of the 32nd ACM International Conference on Multimedia},
  year      = {2024},
  doi       = {10.1145/3664647.3680685}
}

@article{gu2026filopp,
  title   = {FiLo++: Zero-/Few-Shot Anomaly Detection by Fused Fine-Grained Descriptions and Deformable Localization},
  author  = {Gu, Zhaopeng and Zhu, Bingke and Zhu, Guibo and Chen, Yingying and Tang, Ming and Wang, Jinqiao},
  journal = {IEEE Transactions on Circuits and Systems for Video Technology},
  volume  = {36},
  number  = {7},
  pages   = {9616--9630},
  year    = {2026},
  doi     = {10.1109/TCSVT.2026.3671805}
}

@inproceedings{zhu2024inctrl,
  title     = {Toward Generalist Anomaly Detection via In-Context Residual Learning with Few-Shot Sample Prompts},
  author    = {Zhu, Jiawen and Pang, Guansong},
  booktitle = {Proceedings of the IEEE/CVF Conference on Computer Vision and Pattern Recognition},
  pages     = {17826--17836},
  year      = {2024},
  doi       = {10.1109/CVPR52733.2024.01688}
}

@inproceedings{gao2024metauas,
  title     = {MetaUAS: Universal Anomaly Segmentation with One-Prompt Meta-Learning},
  author    = {Gao, Bin-Bin},
  booktitle = {Advances in Neural Information Processing Systems},
  volume    = {37},
  pages     = {39812--39836},
  year      = {2024}
}

@article{gu2024anomalygpt,
  title   = {AnomalyGPT: Detecting Industrial Anomalies Using Large Vision-Language Models},
  author  = {Gu, Zhaopeng and Zhu, Bingke and Zhu, Guibo and Chen, Yingying and Tang, Ming and Wang, Jinqiao},
  journal = {Proceedings of the AAAI Conference on Artificial Intelligence},
  volume  = {38},
  number  = {3},
  pages   = {1932--1940},
  year    = {2024},
  doi     = {10.1609/aaai.v38i3.27963}
}

@inproceedings{jiang2025mmad,
  title     = {MMAD: A Comprehensive Benchmark for Multimodal Large Language Models in Industrial Anomaly Detection},
  author    = {Jiang, Xi and Li, Jian and Deng, Hanqiu and Liu, Yong and Gao, Bin-Bin and Zhou, Yifeng and Li, Jialin and Wang, Chengjie and Zheng, Feng},
  booktitle = {International Conference on Learning Representations},
  year      = {2025}
}

@inproceedings{hu2022lora,
  author       = {Edward J. Hu and
                  Yelong Shen and
                  Phillip Wallis and
                  Zeyuan Allen{-}Zhu and
                  Yuanzhi Li and
                  Shean Wang and
                  Lu Wang and
                  Weizhu Chen},
  title        = {LoRA: Low-Rank Adaptation of Large Language Models},
  booktitle    = {The Tenth International Conference on Learning Representations, {ICLR}
                  2022, Virtual Event, April 25-29, 2022},
  publisher    = {OpenReview.net},
  year         = {2022},
  url          = {https://openreview.net/forum?id=nZeVKeeFYf9},
  bibsource    = {dblp computer science bibliography, https://dblp.org}
}

@inproceedings{meng2024deepstack,
  author       = {Lingchen Meng and
                  Jianwei Yang and
                  Rui Tian and
                  Xiyang Dai and
                  Zuxuan Wu and
                  Jianfeng Gao and
                  Yu{-}Gang Jiang},
  title        = {DeepStack: Deeply Stacking Visual Tokens is Surprisingly Simple and
                  Effective for LMMs},
  booktitle    = {Advances in Neural Information Processing Systems 37: Annual Conference
                  on Neural Information Processing Systems 2024, NeurIPS 2024, Vancouver,
                  BC, Canada, December 10 - 15, 2024},
  year         = {2024},
  url          = {http://papers.nips.cc/paper\_files/paper/2024/hash/29cd7f8331d13ede6dc6d6ef3dfacb70-Abstract-Conference.html},
  bibsource    = {dblp computer science bibliography, https://dblp.org}
}

@article{simeoni2025dinov3,
  author       = {Oriane Sim{\'{e}}oni and
                  Huy V. Vo and
                  Maximilian Seitzer and
                  Federico Baldassarre and
                  Maxime Oquab and
                  Cijo Jose and
                  Vasil Khalidov and
                  Marc Szafraniec and
                  Seung Eun Yi and
                  Micha{\"{e}}l Ramamonjisoa and
                  Francisco Massa and
                  Daniel Haziza and
                  Luca Wehrstedt and
                  Jianyuan Wang and
                  Timoth{\'{e}}e Darcet and
                  Th{\'{e}}o Moutakanni and
                  Leonel Sentana and
                  Claire Roberts and
                  Andrea Vedaldi and
                  Jamie Tolan and
                  John Brandt and
                  Camille Couprie and
                  Julien Mairal and
                  Herv{\'{e}} J{\'{e}}gou and
                  Patrick Labatut and
                  Piotr Bojanowski},
  title        = {DINOv3},
  journal      = {CoRR},
  volume       = {abs/2508.10104},
  year         = {2025},
  url          = {https://doi.org/10.48550/arXiv.2508.10104},
  doi          = {10.48550/ARXIV.2508.10104},
  eprinttype   = {arXiv},
  eprint       = {2508.10104},
  bibsource    = {dblp computer science bibliography, https://dblp.org}
}

@article{tong2024cambrian,
  title={Cambrian-1: A fully open, vision-centric exploration of multimodal llms},
  author={Tong, Shengbang and Brown, Ellis and Wu, Penghao and Woo, Sanghyun and Middepogu, Manoj and Akula, Sai C and Yang, Jihan and Yang, Shusheng and Iyer, Adithya and Pan, Xichen and others},
  journal={Advances in Neural Information Processing Systems},
  volume={37},
  pages={87310--87356},
  year={2024}
}

@inproceedings{DBLP:conf/cvpr/BergmannFSS19,
  author       = {Paul Bergmann and
                  Michael Fauser and
                  David Sattlegger and
                  Carsten Steger},
  title        = {MVTec {AD} - {A} Comprehensive Real-World Dataset for Unsupervised
                  Anomaly Detection},
  booktitle    = {{IEEE} Conference on Computer Vision and Pattern Recognition, {CVPR}
                  2019, Long Beach, CA, USA, June 16-20, 2019},
  pages        = {9592--9600},
  publisher    = {Computer Vision Foundation / {IEEE}},
  year         = {2019},
  url          = {http://openaccess.thecvf.com/content\_CVPR\_2019/html/Bergmann\_MVTec\_AD\_--\_A\_Comprehensive\_Real-World\_Dataset\_for\_Unsupervised\_Anomaly\_CVPR\_2019\_paper.html},
  doi          = {10.1109/CVPR.2019.00982},
  bibsource    = {dblp computer science bibliography, https://dblp.org}
}

@inproceedings{DBLP:conf/visapp/BergmannJSS22,
  author       = {Paul Bergmann and
                  Xin Jin and
                  David Sattlegger and
                  Carsten Steger},
  editor       = {Giovanni Maria Farinella and
                  Petia Radeva and
                  Kadi Bouatouch},
  title        = {The MVTec 3D-AD Dataset for Unsupervised 3D Anomaly Detection and
                  Localization},
  booktitle    = {Proceedings of the 17th International Joint Conference on Computer
                  Vision, Imaging and Computer Graphics Theory and Applications, {VISIGRAPP}
                  2022, Volume 5: VISAPP, Online Streaming, February 6-8, 2022},
  pages        = {202--213},
  publisher    = {{SCITEPRESS}},
  year         = {2022},
  url          = {https://doi.org/10.5220/0010865000003124},
  doi          = {10.5220/0010865000003124},
  bibsource    = {dblp computer science bibliography, https://dblp.org}
}

@inproceedings{DBLP:conf/eccv/ZouJPZD22,
  author       = {Yang Zou and
                  Jongheon Jeong and
                  Latha Pemula and
                  Dongqing Zhang and
                  Onkar Dabeer},
  editor       = {Shai Avidan and
                  Gabriel J. Brostow and
                  Moustapha Ciss{\'{e}} and
                  Giovanni Maria Farinella and
                  Tal Hassner},
  title        = {SPot-the-Difference Self-supervised Pre-training for Anomaly Detection
                  and Segmentation},
  booktitle    = {Computer Vision - {ECCV} 2022 - 17th European Conference, Tel Aviv,
                  Israel, October 23-27, 2022, Proceedings, Part {XXX}},
  series       = {Lecture Notes in Computer Science},
  volume       = {13690},
  pages        = {392--408},
  publisher    = {Springer},
  year         = {2022},
  url          = {https://doi.org/10.1007/978-3-031-20056-4\_23},
  doi          = {10.1007/978-3-031-20056-4\_23},
  bibsource    = {dblp computer science bibliography, https://dblp.org}
}

@inproceedings{DBLP:conf/icumt/JezekJBDS21,
  author       = {Stepan Jezek and
                  Martin Jonak and
                  Radim Burget and
                  Pavel Dvorak and
                  Milos Skotak},
  title        = {Deep learning-based defect detection of metal parts: evaluating current
                  methods in complex conditions},
  booktitle    = {13th International Congress on Ultra Modern Telecommunications and
                  Control Systems and Workshops, {ICUMT} 2021, Brno, Czech Republic,
                  October 25-27, 2021},
  pages        = {66--71},
  publisher    = {{IEEE}},
  year         = {2021},
  url          = {https://doi.org/10.1109/ICUMT54235.2021.9631567},
  doi          = {10.1109/ICUMT54235.2021.9631567},
  bibsource    = {dblp computer science bibliography, https://dblp.org}
}

@article{DBLP:journals/ijcv/BergmannBFSS22,
  author       = {Paul Bergmann and
                  Kilian Batzner and
                  Michael Fauser and
                  David Sattlegger and
                  Carsten Steger},
  title        = {Beyond Dents and Scratches: Logical Constraints in Unsupervised Anomaly
                  Detection and Localization},
  journal      = {Int. J. Comput. Vis.},
  volume       = {130},
  number       = {4},
  pages        = {947--969},
  year         = {2022},
  url          = {https://doi.org/10.1007/s11263-022-01578-9},
  doi          = {10.1007/S11263-022-01578-9},
  bibsource    = {dblp computer science bibliography, https://dblp.org}
}

@article{DBLP:journals/ral/ZhangDBD24,
  author       = {Jian Zhang and
                  Runwei Ding and
                  Miaoju Ban and
                  Linhui Dai},
  title        = {PKU-GoodsAD: {A} Supermarket Goods Dataset for Unsupervised Anomaly
                  Detection and Segmentation},
  journal      = {{IEEE} Robotics Autom. Lett.},
  volume       = {9},
  number       = {3},
  pages        = {2008--2015},
  year         = {2024},
  url          = {https://doi.org/10.1109/LRA.2024.3352358},
  doi          = {10.1109/LRA.2024.3352358},
  bibsource    = {dblp computer science bibliography, https://dblp.org}
}

@inproceedings{DBLP:conf/cvpr/WangZGGZGQCM24,
  author       = {Chengjie Wang and
                  Wenbing Zhu and
                  Bin{-}Bin Gao and
                  Zhenye Gan and
                  Jiangning Zhang and
                  Zhihao Gu and
                  Shuguang Qian and
                  Mingang Chen and
                  Lizhuang Ma},
  title        = {Real-IAD: {A} Real-World Multi-View Dataset for Benchmarking Versatile
                  Industrial Anomaly Detection},
  booktitle    = {{IEEE/CVF} Conference on Computer Vision and Pattern Recognition,
                  {CVPR} 2024, Seattle, WA, USA, June 16-22, 2024},
  pages        = {22883--22892},
  publisher    = {{IEEE}},
  year         = {2024},
  url          = {https://doi.org/10.1109/CVPR52733.2024.02159},
  doi          = {10.1109/CVPR52733.2024.02159},
  bibsource    = {dblp computer science bibliography, https://dblp.org}
}

@inproceedings{DBLP:conf/cvpr/ZhuWZWPZCCGZGWC25,
  author       = {Wenbing Zhu and
                  Lidong Wang and
                  Ziqing Zhou and
                  Chengjie Wang and
                  Yurui Pan and
                  Ruoyi Zhang and
                  Zhuhao Chen and
                  Linjie Cheng and
                  Bin{-}Bin Gao and
                  Jiangning Zhang and
                  Zhenye Gan and
                  Yuxie Wang and
                  Yulong Chen and
                  Shuguang Qian and
                  Mingmin Chi and
                  Bo Peng and
                  Lizhuang Ma},
  title        = {Real-IAD {D3:} {A} Real-World 2D/Pseudo-3D/3D Dataset for Industrial
                  Anomaly Detection},
  booktitle    = {{IEEE/CVF} Conference on Computer Vision and Pattern Recognition,
                  {CVPR} 2025, Nashville, TN, USA, June 11-15, 2025},
  pages        = {15214--15223},
  publisher    = {Computer Vision Foundation / {IEEE}},
  year         = {2025},
  url          = {https://openaccess.thecvf.com/content/CVPR2025/html/Zhu\_Real-IAD\_D3\_A\_Real-World\_2DPseudo-3D3D\_Dataset\_for\_Industrial\_Anomaly\_Detection\_CVPR\_2025\_paper.html},
  doi          = {10.1109/CVPR52734.2025.01417},
  bibsource    = {dblp computer science bibliography, https://dblp.org}
}

@inproceedings{DBLP:conf/cvpr/0007FHDDYP025,
  author       = {Lei Fan and
                  Dongdong Fan and
                  Zhiguang Hu and
                  Yiwen Ding and
                  Donglin Di and
                  Kai Yi and
                  Maurice Pagnucco and
                  Yang Song},
  title        = {{MANTA:} {A} Large-Scale Multi-View and Visual-Text Anomaly Detection
                  Dataset for Tiny Objects},
  booktitle    = {{IEEE/CVF} Conference on Computer Vision and Pattern Recognition,
                  {CVPR} 2025, Nashville, TN, USA, June 11-15, 2025},
  pages        = {25518--25527},
  publisher    = {Computer Vision Foundation / {IEEE}},
  year         = {2025},
  url          = {https://openaccess.thecvf.com/content/CVPR2025/html/Fan\_MANTA\_A\_Large-Scale\_Multi-View\_and\_Visual-Text\_Anomaly\_Detection\_Dataset\_for\_CVPR\_2025\_paper.html},
  doi          = {10.1109/CVPR52734.2025.02376},
  bibsource    = {dblp computer science bibliography, https://dblp.org}
}

@inproceedings{DBLP:conf/iccvw/BaoCLWFWZZ23,
  author       = {Tianpeng Bao and
                  Jiadong Chen and
                  Wei Li and
                  Xiang Wang and
                  Jingjing Fei and
                  Liwei Wu and
                  Rui Zhao and
                  Ye Zheng},
  title        = {{MIAD:} {A} Maintenance Inspection Dataset for Unsupervised Anomaly
                  Detection},
  booktitle    = {{IEEE/CVF} International Conference on Computer Vision, {ICCV} 2023
                  - Workshops, Paris, France, October 2-6, 2023},
  pages        = {993--1002},
  publisher    = {{IEEE}},
  year         = {2023},
  url          = {https://doi.org/10.1109/ICCVW60793.2023.00106},
  doi          = {10.1109/ICCVW60793.2023.00106},
  bibsource    = {dblp computer science bibliography, https://dblp.org}
}

@inproceedings{DBLP:conf/nips/ArodiLBZLPBM24,
  author       = {Akshatha Arodi and
                  Margaux Luck and
                  Jean{-}Luc Bedwani and
                  Aldo Zaimi and
                  Ge Li and
                  Nicolas Pouliot and
                  Julien Beaudry and
                  Ga{\'{e}}tan Marceau{-}Caron},

  title        = {CableInspect-AD: An Expert-Annotated Anomaly Detection Dataset},
  booktitle    = {Advances in Neural Information Processing Systems 37: Annual Conference
                  on Neural Information Processing Systems 2024, NeurIPS 2024, Vancouver,
                  BC, Canada, December 10 - 15, 2024},
  year         = {2024},
  url          = {http://papers.nips.cc/paper\_files/paper/2024/hash/76d9dd096d9469d6b7e732f0cddb51b3-Abstract-Datasets\_and\_Benchmarks\_Track.html},
  bibsource    = {dblp computer science bibliography, https://dblp.org}
}

@inproceedings{DBLP:conf/eccv/ChenLLZ24,
  author       = {Qiyu Chen and
                  Huiyuan Luo and
                  Chengkan Lv and
                  Zhengtao Zhang},
  editor       = {Ales Leonardis and
                  Elisa Ricci and
                  Stefan Roth and
                  Olga Russakovsky and
                  Torsten Sattler and
                  G{\"{u}}l Varol},
  title        = {A Unified Anomaly Synthesis Strategy with Gradient Ascent for Industrial
                  Anomaly Detection and Localization},
  booktitle    = {Computer Vision - {ECCV} 2024 - 18th European Conference, Milan, Italy,
                  September 29-October 4, 2024, Proceedings, Part {LXVII}},
  series       = {Lecture Notes in Computer Science},
  volume       = {15125},
  pages        = {37--54},
  publisher    = {Springer},
  year         = {2024},
  url          = {https://doi.org/10.1007/978-3-031-72855-6\_3},
  doi          = {10.1007/978-3-031-72855-6\_3},
  bibsource    = {dblp computer science bibliography, https://dblp.org}
}

@article{DBLP:journals/corr/abs-2409-06367,
  author       = {Tianwu Lei and
                  Bohan Wang and
                  Silin Chen and
                  Shurong Cao and
                  Ningmu Zou},
  title        = {Texture-AD: An Anomaly Detection Dataset and Benchmark for Real Algorithm
                  Development},
  journal      = {CoRR},
  volume       = {abs/2409.06367},
  year         = {2024},
  url          = {https://doi.org/10.48550/arXiv.2409.06367},
  doi          = {10.48550/ARXIV.2409.06367},
  eprinttype   = {arXiv},
  eprint       = {2409.06367},
  bibsource    = {dblp computer science bibliography, https://dblp.org}
}

@inproceedings{DBLP:conf/aaai/YangXS0ML025,
  author       = {Enquan Yang and
                  Peng Xing and
                  Hanyang Sun and
                  Wenbo Guo and
                  Yuanwei Ma and
                  Zechao Li and
                  Dan Zeng},
  editor       = {Toby Walsh and
                  Julie Shah and
                  Zico Kolter},
  title        = {3CAD: {A} Large-Scale Real-World 3C Product Dataset for Unsupervised
                  Anomaly Detection},
  booktitle    = {Thirty-Ninth {AAAI} Conference on Artificial Intelligence, Thirty-Seventh
                  Conference on Innovative Applications of Artificial Intelligence,
                  Fifteenth Symposium on Educational Advances in Artificial Intelligence,
                  {AAAI} 2025, Philadelphia, PA, USA, February 25 - March 4, 2025},
  pages        = {9175--9183},
  publisher    = {{AAAI} Press},
  year         = {2025},
  url          = {https://doi.org/10.1609/aaai.v39i9.32993},
  doi          = {10.1609/AAAI.V39I9.32993},
  bibsource    = {dblp computer science bibliography, https://dblp.org}
}

@inproceedings{tong2024eyes,
  title={Eyes wide shut? exploring the visual shortcomings of multimodal llms},
  author={Tong, Shengbang and Liu, Zhuang and Zhai, Yuexiang and Ma, Yi and LeCun, Yann and Xie, Saining},
  booktitle={2024 IEEE/CVF Conference on Computer Vision and Pattern Recognition (CVPR)},
  pages={9568--9578},
  year={2024},
  organization={IEEE}
}

@inproceedings{DBLP:conf/cvpr/SalehiSBRR21,
  author       = {Mohammadreza Salehi and
                  Niousha Sadjadi and
                  Soroosh Baselizadeh and
                  Mohammad H. Rohban and
                  Hamid R. Rabiee},
  title        = {Multiresolution Knowledge Distillation for Anomaly Detection},
  booktitle    = {{IEEE} Conference on Computer Vision and Pattern Recognition, {CVPR}
                  2021, virtual, June 19-25, 2021},
  pages        = {14902--14912},
  publisher    = {Computer Vision Foundation / {IEEE}},
  year         = {2021},
  url          = {https://openaccess.thecvf.com/content/CVPR2021/html/Salehi\_Multiresolution\_Knowledge\_Distillation\_for\_Anomaly\_Detection\_CVPR\_2021\_paper.html},
  doi          = {10.1109/CVPR46437.2021.01466},
  bibsource    = {dblp computer science bibliography, https://dblp.org}
}
\bibliographystyle{iclr2027_conference}
\appendix

\section{Implementation and Training Details}
\label{app:vd_implementation}

\subsection{Feature and Injection Configuration}
\label{app:dense_feature_construction}
\label{app:anchored_evidence_details}

Table~\ref{tab:layer_configuration} specifies the feature and injection
configuration for VD-DeepStack-7B. We inject after four approximately
evenly spaced decoder blocks in the first half of the language backbone,
at zero-based indices $\{2,5,9,13\}$.
Appendix~\ref{app:injection_layer_sensitivity} evaluates this choice.
DINO inputs are resized to $16h_i\times16w_i$ for a native patch grid
of $h_i\times w_i$, with non-spatial tokens discarded. Each level uses
a learned softmax mixture of its DINO blocks; the mixing transformations
$O_1,O_2$ and fusion adapters are shared across the image pair.

\begin{table}[htbp]
\centering
\small
\caption{Feature extraction and decoder injection configuration for
VD-DeepStack-7B. All block indices are zero-based.}
\label{tab:layer_configuration}
\setlength{\tabcolsep}{9pt}
\begin{tabular}{@{}cccc@{}}
\toprule
Level & Native visual block & DINO blocks & Decoder block \\
\midrule
1 & 13 & 10--13 & 2 \\
2 & 19 & 14--17 & 5 \\
3 & 25 & 18--20 & 9 \\
4 & 31 & 21--23 & 13 \\
\bottomrule
\end{tabular}
\end{table}

The pair-shared scale $\kappa_k$ uses the detached median
residual-to-native feature norm ratio over reference patches,
with numerical stabilization and a lower bound of $10^{-4}$.
The fusion coefficient is
$\alpha_k=0.01\,\sigma(\theta_k)$.
The matching temperature is initialized to $0.07$ and constrained
to $[0.02,1]$. The channel-wise modulator starts from the identity,
and the spatial readout is
$z_j^k=w_k^\top\operatorname{LN}_k(|Z_j^k|)+b_k$.
Each writer takes the form
$\mathcal T_l^b(x)=B_l^bA_l^b\operatorname{LN}_l^b(x)$,
with a zero-initialized output projection.

\paragraph{Spatial weighting.}
\label{app:decoder_paths}
We normalize scores within each query using the median
$c=\operatorname{median}_t S_t$ and the median absolute deviation
$d=\operatorname{median}_t|S_t-c|$:
\begin{equation}
 a_t=\operatorname{sg}\!\left[
 \sigma\!\left(\operatorname{clip}\!\left(
 \frac{S_t-c}{\max(\rho d,s_{\min})},-b,b\right)\right)\right].
 \label{eq:spatial_authority}
\end{equation}
We use the standard normal-consistency factor $\rho=1.4826$
for MAD scaling, a scale floor $s_{\min}=0.5$ to avoid amplifying
small score fluctuations, and a clipping bound $b=8$ on the sigmoid
input. These weights represent relative spatial emphasis, with
$a_t=1/2$ for a uniform score map. The stop-gradient operator
$\operatorname{sg}$ blocks language gradients through the weights;
the evidence-content path through $E$ remains differentiable.

\paragraph{Residual update limits.}
Context and evidence updates are capped at $\eta_{\rm ctx}=0.0075$
and $\eta_{\rm evi}=0.03$ times the corresponding hidden-state norm.
The cap uses a detached hidden-state norm; evidence is weighted
before clipping, relative to the context-updated query state.
Once an update reaches its cap, increasing its spatial weight no
longer increases the update magnitude.

\subsection{Auxiliary Supervision}
\label{app:supervised_objectives}
\label{app:supervised_schedule}

Query masks are converted to pre-merger targets $M$ by adaptive max
pooling. The evidence objective assigns equal
weight to the mean per-level segmentation loss and the segmentation
loss on mean logits. Each combines Dice loss (smoothing $1$) and mean
binary focal loss (exponent $2$, no class weighting), before pooling.

For the matching objective, let $\mathcal I_N$ and $\mathcal I_A$
denote normal and anomalous query patches. Set
$u_{q,j}^k=\phi_q^k(G_{q,j}^k)$ and
$\widehat u_j^k=\sum_m p_{jm}^k\phi_r^k(G_{r,m}^k)$, and let
$\mu_N^k,\mu_A^k$ be the mean best-match similarities over the two sets.
Then
\begin{equation}
\mathcal L_{\rm match}=\frac1K\sum_k\left[
 \frac1{|\mathcal I_N|}\sum_{j\in\mathcal I_N}
 \bigl(1-\cos(u_{q,j}^k,\widehat u_j^k)\bigr)
 +\max(0,0.1-\mu_N^k+\mu_A^k)\right].
\label{eq:matching_loss_detail}
\end{equation}
Terms involving an empty patch set are omitted.
The shared DINO-mixing transformations use
$\mathcal L_{\rm orth}=d_D^{-2}\sum_{b=1}^{2}
\|O_bO_b^\top-I\|_F^2$, where $d_D$ is the DINO feature dimension.
The joint objective uses $\lambda_m=0.1$ and $\lambda_o=10^{-4}$.
For VD-DeepStack-7B, the coefficient $\lambda_e$ multiplying the complete
evidence loss $\mathcal L_{\mathrm{evi}}$ is set to $1$, as in
Eq.~\ref{eq:joint_loss}.

\subsection{Reinforcement-Learning Refinement}
\label{app:rl_refinement}

GRPO updates only the language backbone from the jointly adapted
checkpoint. We retain 11,000 prompts with nonzero observed reward
variance across four responses sampled from the initial policy.
The reward combines decision correctness, structural validity,
and a localization bonus:
\begin{equation}
r=r_{\mathrm{dec}}+r_{\mathrm{str}}
+0.5\,\mathbb{1}[y=\hat y=1]\,
\max(0,1-0.2N_{\mathrm{miss}}-0.1N_{\mathrm{fp}}).
\label{eq:grpo_reward}
\end{equation}
Here $y=1$ denotes an anomalous image, $\hat y$ is the predicted
decision, and $N_{\mathrm{miss}},N_{\mathrm{fp}}$ count missed and
false-positive boxes. A ground-truth box is covered when any
prediction has IoU$>0.5$. The decision reward $r_{\mathrm{dec}}$ equals $1$ if a Yes/No
decision can be parsed from the answer and agrees with the
ground-truth label, and $0$ otherwise. It is computed independently
of structural validity, so a correct decision can receive this
reward even if the subsequent box format is invalid.

The structural reward $r_{\mathrm{str}}$ equals $1$ only when all
of the following conditions hold: the response follows the
\texttt{<think>...</think><answer>...</answer>} format; the answer
begins with a parseable Yes/No decision; and all subsequent box
objects are fully parseable, each containing a \texttt{bbox\_2d}
field with four integer coordinates and a nonempty string
\texttt{label}. Coordinates must satisfy
$0 \leq x_1 < x_2 \leq W$ and $0 \leq y_1 < y_2 \leq H$,
where $W$ and $H$ denote the image width and height.
In addition, at least one box is required when $y=1$, whereas
no boxes are allowed when $y=0$.
Otherwise, $r_{\mathrm{str}}=0$.
The box-presence requirement depends on the ground-truth label
$y$, rather than the predicted decision $\hat{y}$.
This reward does not assess box overlap with ground truth or
the semantic correctness of defect labels. Training uses 250 steps, four responses
per prompt, temperature $0.9$, KL coefficient $0.04$, and an initial
learning rate of $10^{-6}$ with linear decay.

\section{Supplementary Experiments}
\label{app:design_analyses}

\subsection{Hyperparameter Sensitivity}
\label{app:hyperparameter_sensitivity}

We evaluate the hyperparameter choices for VD-DeepStack-7B on a
validation set of approximately 5,000 examples constructed from MMR-AD.
This sensitivity analysis evaluates models after joint adaptation,
before RL refinement. We report detection accuracy
and localization accuracy at IoU$\geq$0.1.
All scores are percentages; bold marks the best value in each metric
column, and $\dagger$ denotes the final setting.

\paragraph{Injection layers.}
\label{app:injection_layer_sensitivity}
Table~\ref{tab:injection_layer_sensitivity} compares three placements
of four injection sites: over a wider depth range, within the first half,
or within the first quarter of the decoder.
The first-half placement used in the final 7B model yields the highest
detection accuracy. The first-quarter placement gives the highest
localization accuracy, 0.24 percentage points above the first-half setting.

\begin{table}[t]
\centering
\caption{Sensitivity to decoder injection layers (zero-based indices).}
\label{tab:injection_layer_sensitivity}
\small
\setlength{\tabcolsep}{2pt}
\renewcommand{\arraystretch}{1.12}
\begin{tabular}{@{}lccc@{}}
\toprule
Placement & Blocks & Det. Acc. & Loc. Acc. \\
\midrule
Wide span & 5/11/16/22 & 78.54 & 43.74 \\
First half$^\dagger$ & 2/5/9/13 & \textbf{79.65} & 44.47 \\
First quarter & 0/2/4/6 & 79.30 & \textbf{44.71} \\
\bottomrule
\end{tabular}
\end{table}

\subsection{Frozen Visual-Encoder Comparison}
\label{app:dino_local_features}

\begin{figure}[htbp]
\centering
\includegraphics[width=\linewidth]{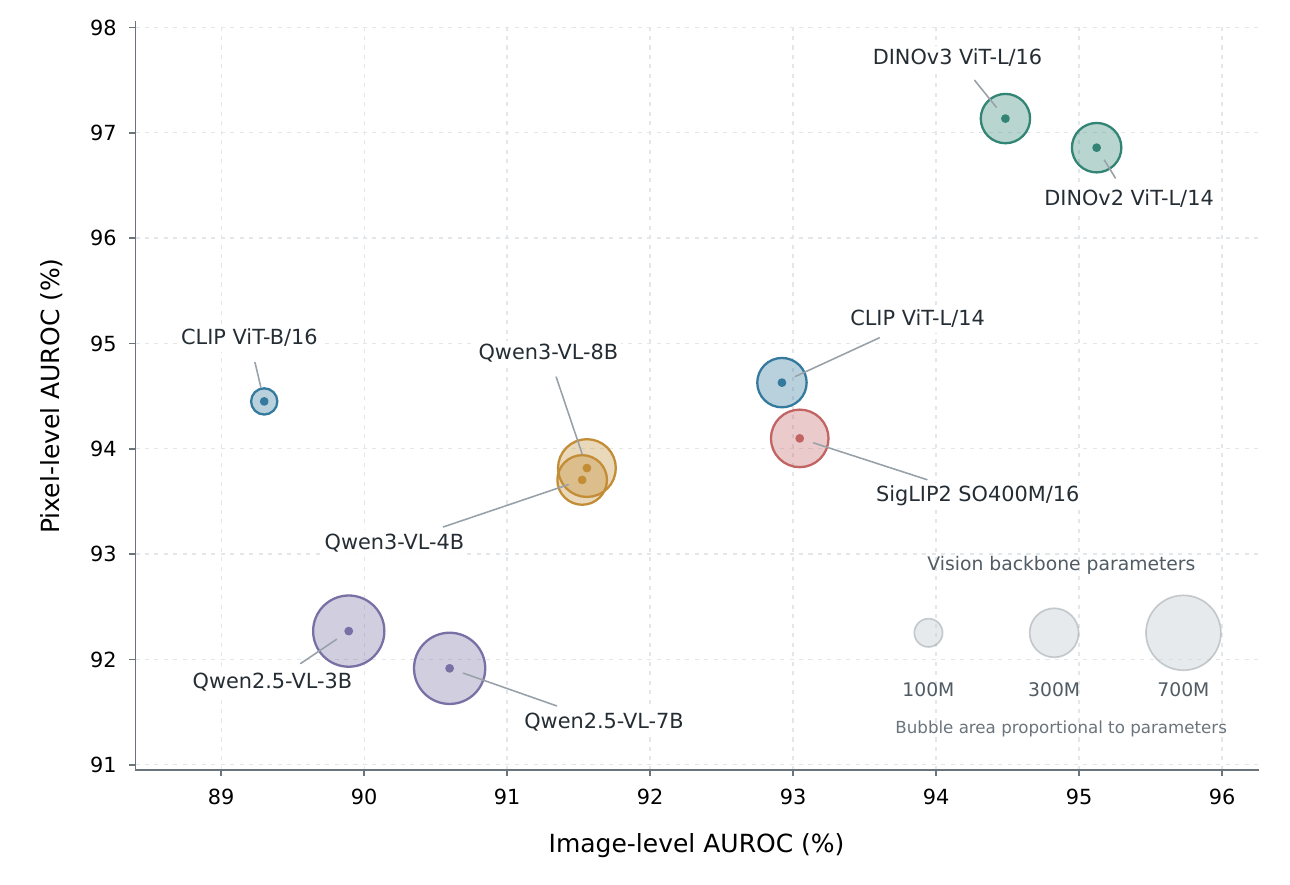}
\caption{Frozen visual encoders under a common one-shot matching
protocol. Bubble area indicates vision-backbone parameter count;
Qwen labels identify the parent vision--language models.}
\label{fig:encoder_comparison}
\end{figure}

We compare nine frozen visual encoders on the 15 MVTec AD categories.
All encoders use the same one normal reference per category, excluded
from evaluation queries. Inputs are $336\times336$, except for
Qwen3-VL ($352\times352$). We extract patch features from four
uniformly sampled layers in the latter half of each encoder.
At each layer, every query patch is matched to its nearest reference
patch by cosine similarity. One minus the mean similarity across
layers gives the anomaly map; its maximum gives the image score.
Image-level and pixel-level AUROC are averaged equally across categories.

Figure~\ref{fig:encoder_comparison} shows higher image-level and
pixel-level AUROC for DINOv3 ViT-L/16 than for the native visual encoder
of Qwen2.5-VL-7B. This comparison motivates DINOv3 as a source of
fine-grained features for query--reference matching; it does not
isolate the effect of fusion in the complete model.

\subsection{Localization at a Stricter IoU Threshold}
\label{app:strict_localization}

Table~\ref{tab:loc_iou05} reports localization at IoU$\geq$0.5,
complementing the IoU$\geq$0.1 evaluation in the main text.
The stricter threshold requires closer agreement between predicted
and ground-truth boxes.

\begin{table*}[t]
\centering
\small
\caption{
Localization results (\%) at IoU$\geq$0.5.
}
\label{tab:loc_iou05}
\setlength{\tabcolsep}{3.0pt}
\renewcommand{\arraystretch}{1.12}
\resizebox{\textwidth}{!}{%
\begin{tabular}{@{}l*{12}{c}@{}}
\toprule
\multirow{2}{*}{\textbf{Model}}
& \multicolumn{3}{c}{\textbf{MVTec AD}}
& \multicolumn{3}{c}{\textbf{VisA}}
& \multicolumn{3}{c}{\textbf{MVTec 3D}}
& \multicolumn{3}{c}{\textbf{MPDD}} \\
\cmidrule(lr){2-4}
\cmidrule(lr){5-7}
\cmidrule(lr){8-10}
\cmidrule(lr){11-13}
& Acc. & Recall & Prec.
& Acc. & Recall & Prec.
& Acc. & Recall & Prec.
& Acc. & Recall & Prec. \\
\midrule

Qwen3.8-Flash
& 18.9 & 29.4 & 30.4
& 6.4 & 11.0 & 13.0
& 7.9 & 13.0 & 14.4
& 10.9 & 17.5 & 18.4 \\

Qwen3-VL-32B
& 29.6 & 45.7 & 41.9
& 11.0 & 16.4 & 22.8
& 16.1 & 25.4 & 28.7
& 26.5 & 38.1 & 32.2 \\

Qwen3.8-27B
& 41.7 & 53.7 & 57.3
& 13.7 & 21.6 & 26.6
& 18.7 & 28.2 & 32.4
& 25.8 & 36.7 & 32.0 \\

IAD-R1*
& 34.9 & 44.3 & 55.2
& 24.2 & 29.3 & 53.5
& 26.0 & 34.6 & 52.0
& 23.3 & 27.5 & 40.6 \\

\midrule
\textbf{VD-DeepStack-7B}
& 42.6 & 53.4 & 61.7
& 26.1 & 32.7 & 52.0
& 35.0 & 42.3 & 60.9
& 30.2 & 36.6 & 48.3 \\

\textbf{VD-DeepStack-8B}
& 49.2 & 59.2 & 70.2
& 30.4 & 37.1 & 58.6
& 43.2 & 49.0 & 70.5
& 34.4 & 39.1 & 50.9 \\

\bottomrule
\end{tabular}%
}
\end{table*}
\subsection{CoT Evaluation Protocol}
\label{app:llm_judge_protocol}

GLM-5.3-Flash compares generated CoTs with ground-truth annotations
using text alone, with model identities withheld and final answers
excluded. Integer scores from 0 to 100 reflect the anomaly judgment
expressed in the CoT, defect type, location, and fabricated claims;
neither writing style nor length is rewarded.
The scoring anchors in Table~\ref{tab:llm_judge_rubric} are selected
by the ground-truth label $y$.
Location agreement is assessed against GT coordinates or spatial
descriptions.
The Overall score averages sample-level scores across all evaluated
samples from the four datasets.

\begin{table}[t]
\centering
\caption{Scoring anchors for generated CoTs.
GT denotes the ground-truth annotation.}
\label{tab:llm_judge_rubric}
\small
\setlength{\tabcolsep}{4pt}
\renewcommand{\arraystretch}{1.15}
\begin{tabular}{@{}p{0.16\linewidth}p{0.13\linewidth}p{\dimexpr0.71\linewidth-4\tabcolsep\relax}@{}}
\toprule
Sample & Score & Criterion \\
\midrule
Abnormal\newline($y=1$) & 90--100 & Correct anomaly judgment, with defect type and location consistent with GT. \\
& 75--89 & Correct type with a vague or slightly inaccurate location, or correct location with a synonymous or near-equivalent type. \\
& 55--74 & Correct broad region but incorrect type, or only one of multiple GT defects identified. \\
& 30--54 & Anomaly stated, but both the described type and location differ from GT. \\
& 10--29 & Normal judgment or missed anomaly, or irrelevant content. \\
& 0--9 & Empty CoT or wholly fabricated content. \\
\midrule
Normal\newline($y=0$) & 90--100 & Correct normal judgment with an explanation consistent with GT. \\
& 75--89 & Correct normal judgment with minor explanation errors, such as misidentifying a component. \\
& 55--74 & Correct normal judgment with substantial incorrect statements. \\
& 46--54 & Severe indecision or pervasive contradictions. \\
& 30--45 & Tentative false positive, such as describing a ``possible defect.'' \\
& 10--29 & Confident false positive; fabricated coordinates place the score nearer the lower bound. \\
& 0--9 & Empty or incoherent CoT. \\
\bottomrule
\end{tabular}
\end{table}

\clearpage
\subsection{Additional Qualitative Comparisons}
\label{app:additional_qualitative}

\begin{figure}[htbp]
\centering
\includegraphics[width=\linewidth]{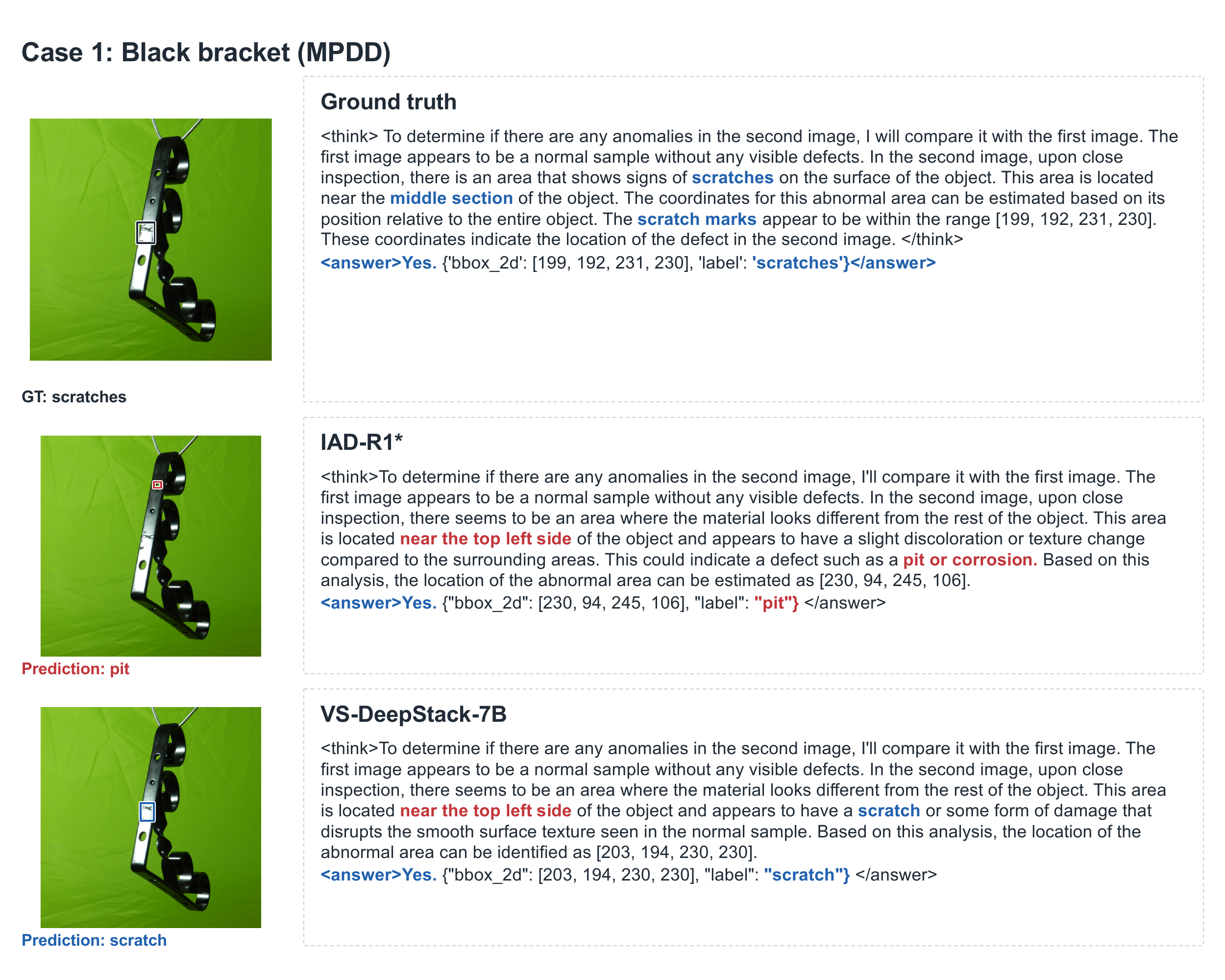}
\caption{An anomalous black bracket from MPDD, showing GT annotations
and the two models' responses and predicted boxes. Red keywords
disagree with GT; blue keywords convey similar or matching content.}
\label{fig:app_case_mpdd}
\end{figure}

\begin{figure}[htbp]
\centering
\includegraphics[width=\linewidth]{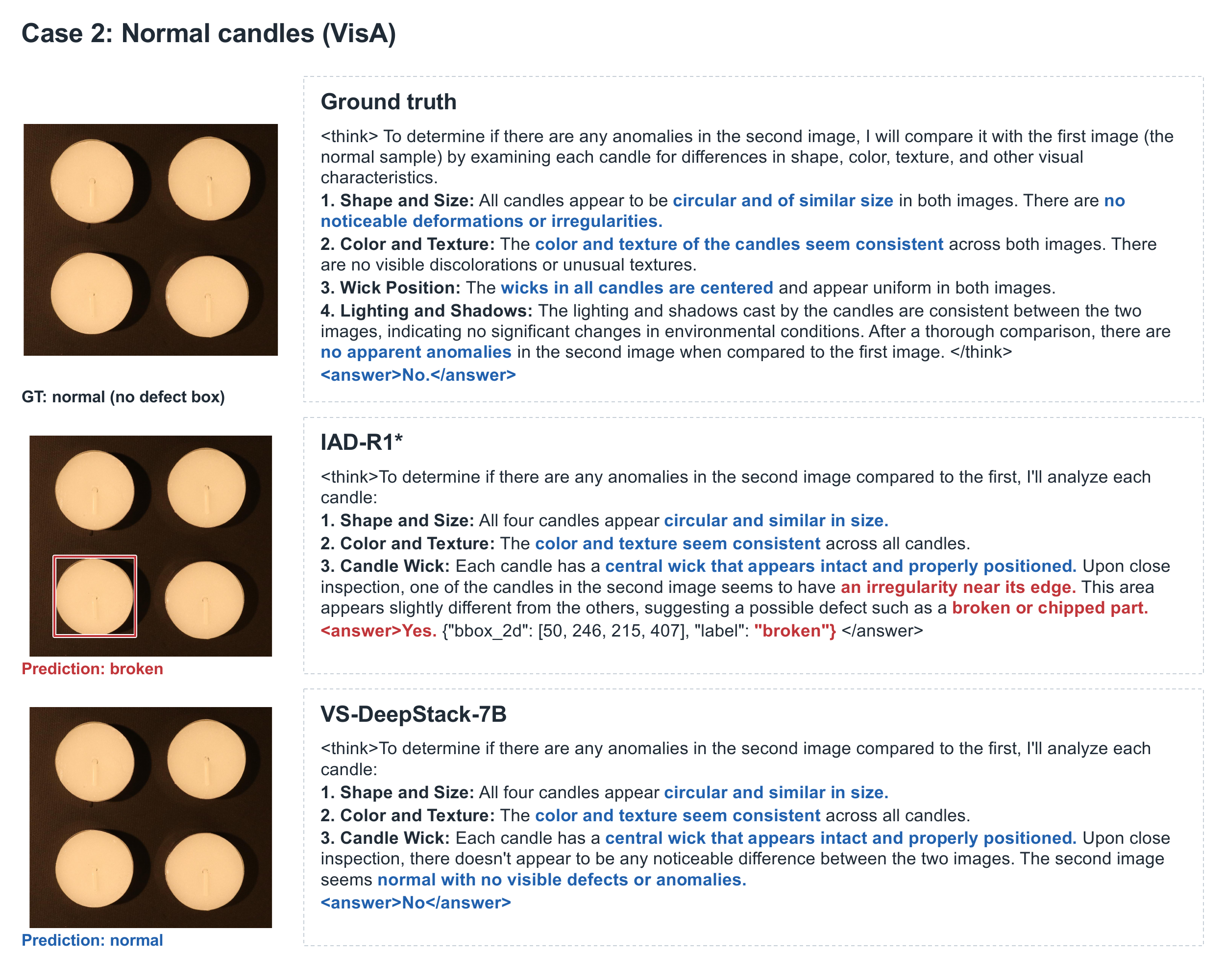}
\caption{A normal candle sample from VisA. IAD-R1$^*$ reports a broken
part, while VD-DeepStack-7B predicts no anomaly. Red keywords disagree
with GT; blue keywords convey similar or matching content.}
\label{fig:app_case_visa}
\end{figure}

\begin{figure}[htbp]
\centering
\includegraphics[width=\linewidth]{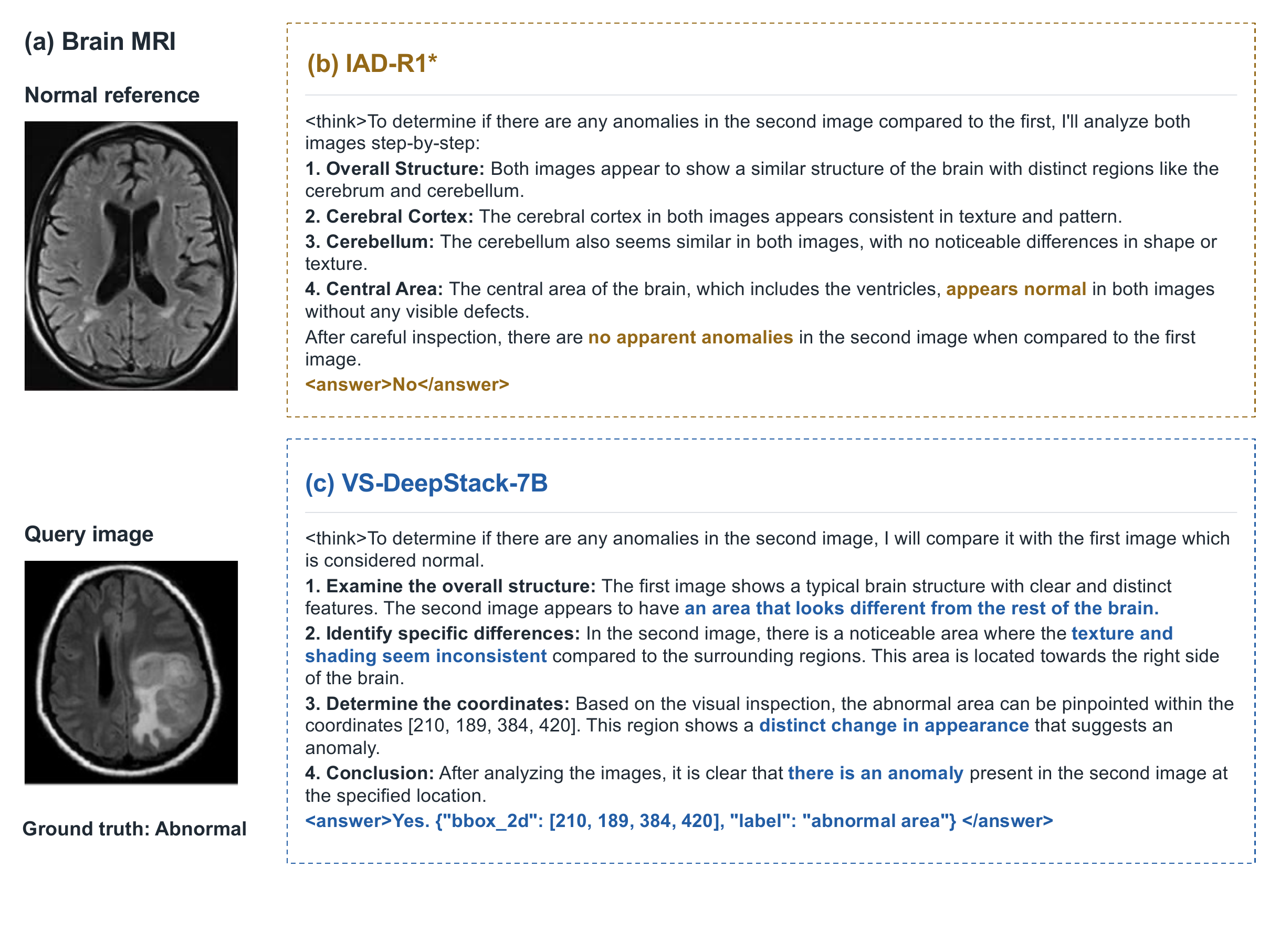}
\caption{An abnormal BrainMRI query paired with a normal reference.
Blue highlights mark the response identifying an anomaly, and ochre
highlights mark the incorrect normal judgment. GT supports the
image-level decision only; the highlighted regional descriptions and
coordinates are unverified model outputs.}
\label{fig:app_case_brainmri}
\end{figure}

Figures~\ref{fig:app_case_mpdd}--\ref{fig:app_case_brainmri}
provide additional comparisons between VD-DeepStack-7B and IAD-R1$^*$
on abnormal and normal industrial samples and a medical image.

In the MPDD example (Figure~\ref{fig:app_case_mpdd}), both models
predict an anomaly. IAD-R1$^*$ describes a pit or corrosion and places
its box away from the annotated scratch. VD-DeepStack-7B identifies
a scratch and predicts a box close to GT. Its CoT still contains the
incorrect phrase ``near the top left side,'' illustrating a remaining
inconsistency between the verbal location and the predicted box.

\clearpage
For the normal VisA sample (Figure~\ref{fig:app_case_visa}), both
models initially describe consistent candle shapes, textures, and
intact wicks. IAD-R1$^*$ then introduces an unsupported claim of a
broken or chipped edge and outputs a defect box. VD-DeepStack-7B
maintains a normal judgment consistent with GT, avoiding this false
positive.

\clearpage
In the BrainMRI example (Figure~\ref{fig:app_case_brainmri}), IAD-R1$^*$
emphasizes similarities to the normal reference and incorrectly
classifies the query as normal. VD-DeepStack-7B describes a region
with different texture and shading and predicts an anomaly, matching
the image-level label. The regional description and generated box
remain unverified because spatial annotations are unavailable.

\end{document}